\pdfoutput=1

\documentclass[11pt,table,dvipsnames]{article}

\usepackage{acl}
\usepackage{times}
\usepackage{latexsym}
\usepackage[T1]{fontenc}

\usepackage[utf8]{inputenc}

\usepackage{microtype}
\usepackage{latexsym}
\usepackage{dsfont}
\usepackage{booktabs} 
\usepackage{subfigure}
\usepackage{CJKutf8}
\usepackage{graphicx}
\usepackage{bigstrut}
\usepackage{multirow}
\usepackage{amsmath}
\usepackage{multirow, makecell, caption}
\usepackage{floatrow}
\newfloatcommand{capbtabbox}{table}[][\FBwidth]

\usepackage[normalem]{ulem}
\usepackage{amssymb}
\usepackage{amsfonts}
\usepackage{multicol}
\usepackage{tipa}
\usepackage{amsmath}
\usepackage{bm}
\usepackage[ruled,linesnumbered]{algorithm2e}

\usepackage{paralist}
\usepackage[switch]{lineno}
\usepackage{enumitem}
\usepackage{multirow, makecell, caption}
\usepackage{arydshln}
\usepackage{verbatim}
\usepackage[normalem]{ulem}
\usepackage{amssymb}
\usepackage{amsfonts}
\usepackage{multicol}

\usepackage{csquotes}
\usepackage{xspace}
\usepackage{paralist}
\usepackage{mdwlist}
\usepackage{subfigure}
\usepackage{makecell}
\usepackage{array}
\usepackage{enumitem}
\usepackage{bm}
\usepackage{colortbl}
\usepackage{dsfont}
\usepackage{url}
\usepackage{booktabs} 
\usepackage{graphicx}
\usepackage{tabularx}
\usepackage{amsmath}
\usepackage{bbm}
\usepackage{tikz}
\usepackage{pgfplots}
\newcommand{\method}{\texttt{CoScript}\xspace}
\newcommand{\proScript}{\texttt{proScript}\xspace}
\newcommand{\wikiHow}{\texttt{wikiHow}\xspace}
\newcommand{\todo}{$\blacksquare$ TODO\xspace}
\newcommand{\ie}{\textit{i.e.}\xspace}
\newcommand{\eg}{\textit{e.g.}\xspace}
\newcommand{\wo}{\textit{w/o}\xspace}

\makeatletter
\def\adl@drawiv#1#2#3{%
        \hskip.5\tabcolsep
        \xleaders#3{#2.5\@tempdimb #1{1}#2.5\@tempdimb}%
                #2\z@ plus1fil minus1fil\relax
        \hskip.5\tabcolsep}
\newcommand{\cdashlinelr}[1]{%
  \noalign{\vskip\aboverulesep
           \global\let\@dashdrawstore\adl@draw
           \global\let\adl@draw\adl@drawiv}
  \cdashline{#1}
  \noalign{\global\let\adl@draw\@dashdrawstore
           \vskip\belowrulesep}}
\makeatother

\author{Siyu Yuan\textsuperscript{\rm $\heartsuit$},
 Jiangjie Chen\textsuperscript{\rm $\spadesuit$}\thanks{~~Corresponding authors.}, 
 Ziquan Fu\textsuperscript{\rm $\clubsuit$}\thanks{~~Work done while at Brain Technologies, Inc.}, 
 Xuyang Ge\textsuperscript{\rm $\spadesuit$},\\
 \bf Soham Shah\textsuperscript{\rm $\diamondsuit$}, 
 Charles Robert Jankowski\textsuperscript{\rm $\diamondsuit$}, 
 Yanghua Xiao\textsuperscript{\rm $\spadesuit\P$}, 
 Deqing Yang\textsuperscript{\rm $\heartsuit$}\footnotemark[1]\\
\textsuperscript{\rm $\heartsuit$}School of Data Science, Fudan University\\
\textsuperscript{\rm $\spadesuit$}Shanghai Key Laboratory of Data Science, School of Computer Science, Fudan University\\
\textsuperscript{\rm $\clubsuit$}System Inc. 
\textsuperscript{\rm $\diamondsuit$}Brain Technologies, Inc. \\
\textsuperscript{\rm $\P$}Fudan-Aishu Cognitive Intelligence Joint Research Center\\
\texttt{syyuan21@m.fudan.edu.cn, jjchen19@fudan.edu.cn} \\\texttt{frank@system.com, \{sshah,cjankowski\}@brain.im} \\ \texttt{\{xyge20,shawyh,yangdeqing\}@fudan.edu.cn}
}

\title{ 
Distilling Script Knowledge from Large Language Models for \\Constrained Language Planning
}

\begin{document}

\maketitle

\begin{abstract}

In everyday life, humans often plan their actions by following step-by-step instructions in the form of goal-oriented scripts. 
Previous work has exploited language models (LMs) to plan for abstract goals of stereotypical activities (\eg, ``\textit{make a cake}''), but leaves more specific goals with multi-facet constraints understudied (\eg, ``\textit{make a cake for diabetics}'').
In this paper, we define the task of constrained language planning for the first time.
We propose an over-generate-then-filter approach to improve large language models (LLMs) on this task, and use it to distill a novel constrained language planning dataset, \method, which consists of 55,000 scripts.
Empirical results demonstrate that our method significantly improves the constrained language planning ability of LLMs, especially on constraint faithfulness.
Furthermore, \method is demonstrated to be quite effective in endowing smaller LMs with constrained language planning ability. \footnote{Resources of this paper can be found at \url{https://github.com/siyuyuan/coscript}.}
\end{abstract}

\section{Introduction}
\label{sec:intro}

\begin{figure}[t]
    \centering
    \includegraphics[width=\linewidth]{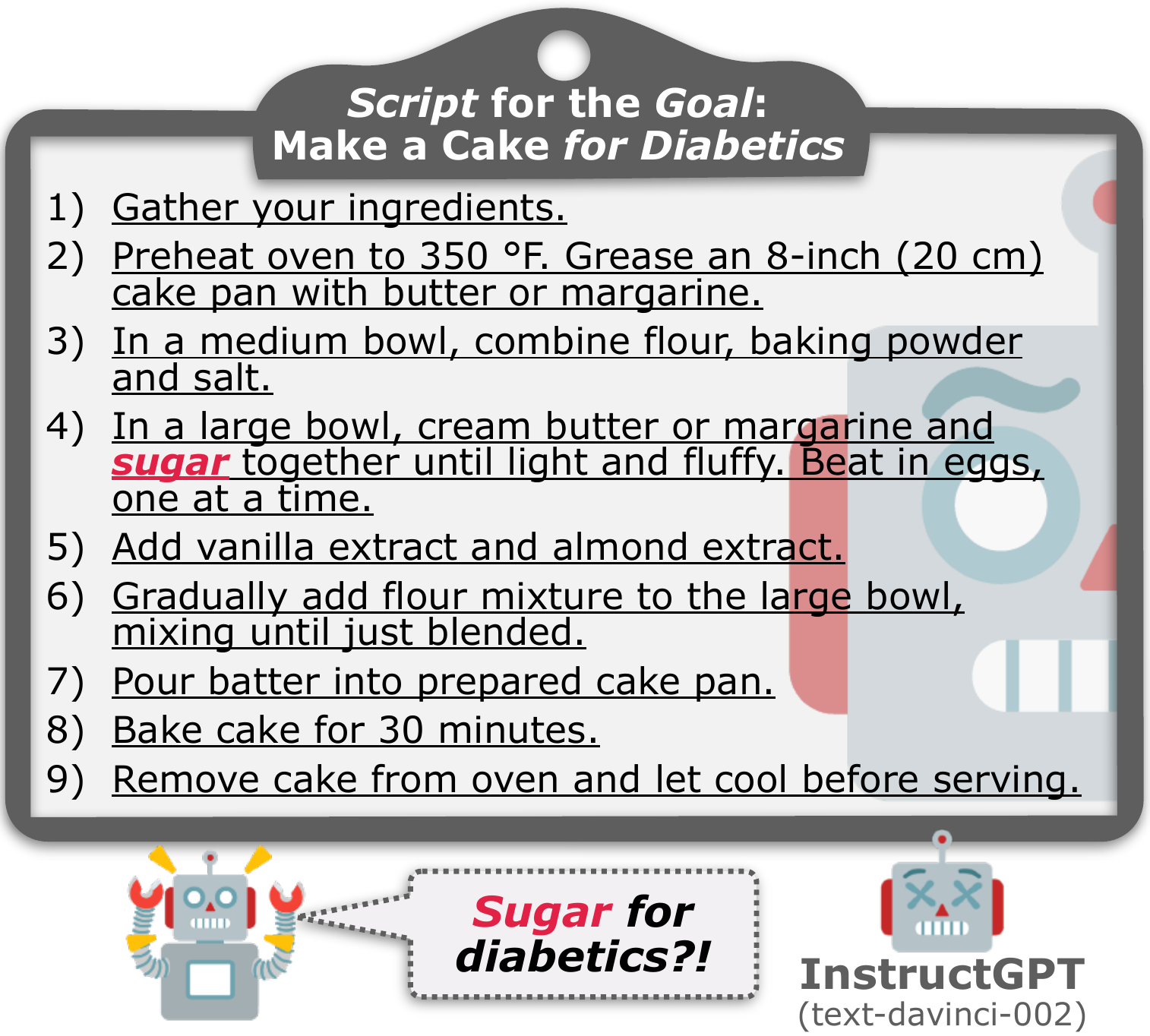}
    \caption{A list of steps InstructGPT generates to plan for the goal ``\textit{make a cake for diabetics}''. 
    InstructGPT mistakenly adds sugar to the cake, which is unfit for diabetic patients.
    This example shows that InstructGPT sometimes cannot effectively and faithfully script for a \textit{specific} goal with fine-grained constraints.}
    \label{fig:example}
\end{figure}

To accomplish everyday goals, humans usually plan their actions in accordance with step-by-step instructions.
Such instructions are discovered as \textit{goal-oriented scripts}~\cite{DBLP:conf/ijcai/SchenkA75,schank2013scripts}, involving a set of prototypical event sequences to achieve goals.
For the example in Figure~\ref{fig:example}, to achieve the goal (\textit{make a cake}), one usually has to follow certain steps of instructions, \eg, \textit{gather ingredients}, \textit{preheat the oven}, etc.
The planning for such step-by-step scripts chains up reasoning toward complex goals~\cite{abelson1976script,wei2022chain}.
Therefore, the automation of planning envisions more intelligent and reasonable AI systems in various domains, such as executable robotic systems~\cite{Kovalchuk_Shekhar_Brafman_2021,huang2022language} and reasoning systems for problem-solving~\cite{wei2022chain,wang2022self}.

Recent studies have identified that language models (LMs) can be used to plan scripts~\cite{sancheti-rudinger-2022-large}.
Previous work~\cite{huang2022language} has shown that large language models (LLMs), such as GPT-3~\cite{NEURIPS2020_1457c0d6}  InstructGPT~\cite{ouyang2022training} and PaLM~\cite{chowdhery2022palm}, can effectively decompose goals into procedural steps in a zero-/few-shot manner.
To train specialized models, researchers have proposed datasets for the automatic understanding and generation of script knowledge~\cite{DBLP:conf/ijcai/SchenkA75,regneri-etal-2010-learning,wanzare-etal-2016-crowdsourced,lyu-etal-2021-goal,sakaguchi-etal-2021-proscript-partially}.
However, previous work mainly focuses on planning for the abstract goals of stereotypical activities~\cite{abelson1976script}.
Planning for goals with specific constraints (\eg, \textit{for diabetics}) still remains under-studied.

In this paper, we define the problem of \textit{constrained language planning}, which imposes different constraints on the goals of planning.
An \textit{abstract goal}, for example, \textit{make a cake}, can be inherited by different real-life \textit{specific goals} with multi-faceted \textit{constraints}.
A cake can be made for 
\begin{inparaenum}[\it 1)]
    \item different ingredients (\eg, \textit{chocolate} or \textit{vanilla});
    \item various tools (\eg, with a \textit{microwave} or an \textit{oven}); or
    \item different purposes (\eg, for a \textit{wedding} or a \textit{birthday party}).
\end{inparaenum}
A good planner should write scripts that are reasonable and faithful to constraints. 
However, LLMs sometimes do not plan faithfully toward the constraints.
As showcased in Figure~\ref{fig:example}, InstructGPT suggests adding sugar to the cake for diabetic patients.
Also, due to a shortage of datasets for constrained language planning, the ability of smaller but specialized models to plan with specific constraints has been underexplored.

In this paper, we aim to evaluate and improve the constrained language planning ability of LLMs, while distilling a dataset from LLMs to train specialized models.
Our empirical study finds that LLMs tend to plan fluently but unfaithfully to the constraints.
Thus, we employ an over-generate-then-filter approach~\cite{wiegreffe-etal-2022-reframing} to satisfy the quality of the generated scripts to constraints.
The main idea is to select high-quality ones from multiple generated scripts.
Then, we use LLMs (\eg, InstructGPT) with this approach to generate a dataset for constrained language planning, which inherits the idea of symbolic knowledge distillation from models~\cite{west-etal-2022-symbolic}.
We thus arrive at a \textbf{Co}nstrained \textbf{Script} dataset, \ie, \method, which consists of 55,000 high-quality scripts with specific goals and steps. 
Experiments show that, when trained on \method, smaller models such as  T5~\cite{raffel2020exploring} can achieve good performance, even surpassing that of LLMs.

Our contributions are summarized as follows:
\begin{inparaenum}[\it 1)]
    \item To our knowledge, we are the first to establish the constrained language planning problem, which advances language planning toward more specific goals.
    \item We evaluate the few-shot constrained language planning ability of LLMs and develop an over-generate-then-filter method for LLMs, resulting in a 26\% increase in accuracy.
    \item Based on our method, we use LLMs to generate a high-quality script dataset (\method) for constrained language planning. 
    By leveraging the \method, we endow specialized and smaller models with constrained language planning ability, which achieves comparable performance to that of LLMs.
\end{inparaenum}

\section{Related Work}
\label{sec:related}

\paragraph{Language Planning}
Language planning aims to decompose a goal into sequences of steps~\cite{kaplan1997language}, which is widely used in robotics~\cite{6907402,8794441,9812355} and procedural text generation~\cite{goldfarb-tarrant-etal-2020-content,hu-etal-2022-planet}.
Early studies approach language planning with syntactic parsing for the context~\cite{koller-stone-2007-sentence,garoufi-koller-2010-automated}.
Recently, researchers have investigated the planning capability of language models in various domains~\cite{olmo2021gpt3,valmeekam2022large}.
However, they mainly focus on generating scripts for stereotypical activities toward abstract goals.
For example, \citet{huang2022language} proposes to plan for the general-typed tasks for embodied agents, while \citet{yang2021induce} edits actions for abstract goals to video retrieval.
In contrast, we explore planning for specific goals (\eg, ``\textit{make a cake for diabetics}'').
\citet{collins2022structured} has benchmarked LLMs for planning with included/excluded objects, but they merely study this problem in a limited scope (only dozens of cases) without further in-depth analysis.

\paragraph{Scripts}
A structure describing a sequence of events in a particular scenario is \textit{script}~\cite{DBLP:conf/ijcai/SchenkA75}, consisting of two types:
\begin{inparaenum}[\it 1)]
    \item \textit{Narrative script}: 
    a narrative chain of events describing a particular scenario derived from narrative texts such as recipes~\cite{fang-etal-2022-take} or stories~\cite{tandon-etal-2020-dataset};
    \item \textit{Goal-oriented script}~\cite{regneri-etal-2010-learning,wanzare-etal-2016-crowdsourced}: an appropriate sequence of steps as instructions to achieve a goal.
\end{inparaenum}
In this work, the steps for achieving a given goal in language planning can be categorized into the second class.
Many datasets for goal-oriented scripts have been proposed to improve the language planning ability of LMs~\cite{sakaguchi-etal-2021-proscript-partially,lyu-etal-2021-goal}.
However, they mainly consist of abstract goals with prototypical instructions and thus are not built to train LMs for planning with more specific goals.

\paragraph{In-Context Learning}
With the great success of LLMs~\cite{NEURIPS2020_1457c0d6,ouyang2022training,chowdhery2022palm}, \textit{in-context learning}~\cite{NEURIPS2020_1457c0d6,min2022rethinking} has established its great task-solving potentials with a textual task instruction and a few examples.
Moreover, when being used for dataset construction, the data samples that LLMs generate can sometimes outperform crowd-sourced human-authored data in factuality and fluency~\cite{lu-etal-2022-fantastically,min2022rethinking}.
This shows a promising alternative to costly large-scale crowd-sourcing to construct datasets using LLMs~\cite{wiegreffe-etal-2022-reframing,liu2022wanli,west-etal-2022-symbolic}.
Inspired by these studies, in our work, we adopt the in-context learning for LLMs not only for better language planning, but also as a reliable \textit{crowd-worker} to scale up the planning data into a reusable dataset to train smaller models.

\section{Definitions}
\label{sec:Definition}

\begin{table}[t]
\small
    \centering
    \begin{tabularx}{\linewidth}{|X|}
    \hline
        \makecell[c]{\textbf{Constraint Type 1: \textit{Modifier}}} \\
        \hline
        \textbf{Definition}: A word, an adjective or a phrase that modifies or constrains an abstract goal. \\
        \hline
        \textbf{Ex.1}: Make a \colorbox[rgb]{0.99, 0.95, 0.93}{\color[rgb]{0.81, 0.41, 0.22}chocolate} cake.\\
        \textbf{Ex.2}: Make a \colorbox[rgb]{0.99, 0.95, 0.93}{\color[rgb]{0.81, 0.41, 0.22}pink} cake. \\
        \hline
        \hline
        \makecell[c]{\textbf{Constraint Type 2: \textit{Method}}} \\
        \hline
        \textbf{Definition}: A tool or specified mode that controls the process for achieving the goal. \\
        \hline
        \textbf{Ex.1}: Make a cake \colorbox[rgb]{0.99, 0.95, 0.93}{\color[rgb]{0.81, 0.41, 0.22}with an oven}. \\
        \textbf{Ex.2}: Make a cake \colorbox[rgb]{0.99, 0.95, 0.93}{\color[rgb]{0.81, 0.41, 0.22}by using cake mix}. \\
        \hline
        \hline
        \makecell[c]{\textbf{Constraint Type 3: \textit{Intent}}}\\
        \hline
        \textbf{Definition}: An additional purpose or demand when completing the goal. \\
        \hline
        \textbf{Ex.1}: Make a cake \colorbox[rgb]{0.99, 0.95, 0.93}{\color[rgb]{0.81, 0.41, 0.22}for wedding}. \\
        \textbf{Ex.2}: Make a cake \colorbox[rgb]{0.99, 0.95, 0.93}{\color[rgb]{0.81, 0.41, 0.22}for diabetics}. \\
    \hline
    \end{tabularx}
    \caption{Three types of constraints and their definitions that are used to prompt for new instances of specific goals.
    In the examples (\textbf{Ex.}), upon the abstract goal, we give two instances for each type of constraint by combining the goal with constraints into specific goals.
    The constraint within each example is  \colorbox[rgb]{0.99, 0.95, 0.93}{\color[rgb]{0.81, 0.41, 0.22}highlighted}.
    }
    \label{tb:factors}
\end{table}

Before diving into technical details, we first clarify some important terms used in the paper.

\paragraph{Scripts} 
A goal-oriented \textit{script} is \textit{a list of steps} ($\mathbf{S} = \{s_1, s_2, \cdots, s_{|\mathbf{S}|}\}$) that fulfill a certain \textit{goal} ($\mathcal{G}$) (\eg, ``\textit{make a cake}'')~\cite{suddendorf2007evolution,schank2013scripts}.
The language planning task is defined as $\mathcal{M}: \mathcal{G}\rightarrow \mathbf{S}$, where $\mathcal{M}$ is the planning model.
\paragraph{Goals} 
Different from previous studies that focus mostly on abstract goals with prototypical scripts, we define a taxonomic structure of goals by extending the derivatives of abstract goals.
We define a \textit{specific goal} that inherits from an \textit{abstract one} but with new information as a constraint to limit the scope.
An \textbf{\textit{abstract goal}}, denoted as $\mathcal{G}_{a}$, refers to stereotypical activities, \eg, ``\textit{make a cake}''.
A \textbf{\textit{specific goal}}, denoted as $\mathcal{G}_{c}$, is derived from the corresponding $\mathcal{G}_a$ with various constraints, \eg, ``\textit{make a \underline{chocolate} cake}''.

\paragraph{Constraints and Constrained Language Planning}
To enrich the semantics of specific goals, we 
define three types of \textbf{\textit{constraints}}, \ie, \textit{modifier}, \textit{method} and \textit{intent}, as shown in Table~\ref{tb:factors}. 
They express different angles of extending an abstract goal and can be further instantiated and concreted.
\textbf{\textit{Constrained language planning}} denotes generating a constraint-faithful script $\mathbf{S}: \mathbf{S} = \mathcal{M}(\mathcal{G}_c)$ toward specific goals ($\mathcal{G}_c$) with various constraints ($\mathcal{C}$).

\section{Constrained Language Planning with LLMs}
\label{sec:LLM}

\begin{figure}[t]
    \centering
    \includegraphics[width=\linewidth]{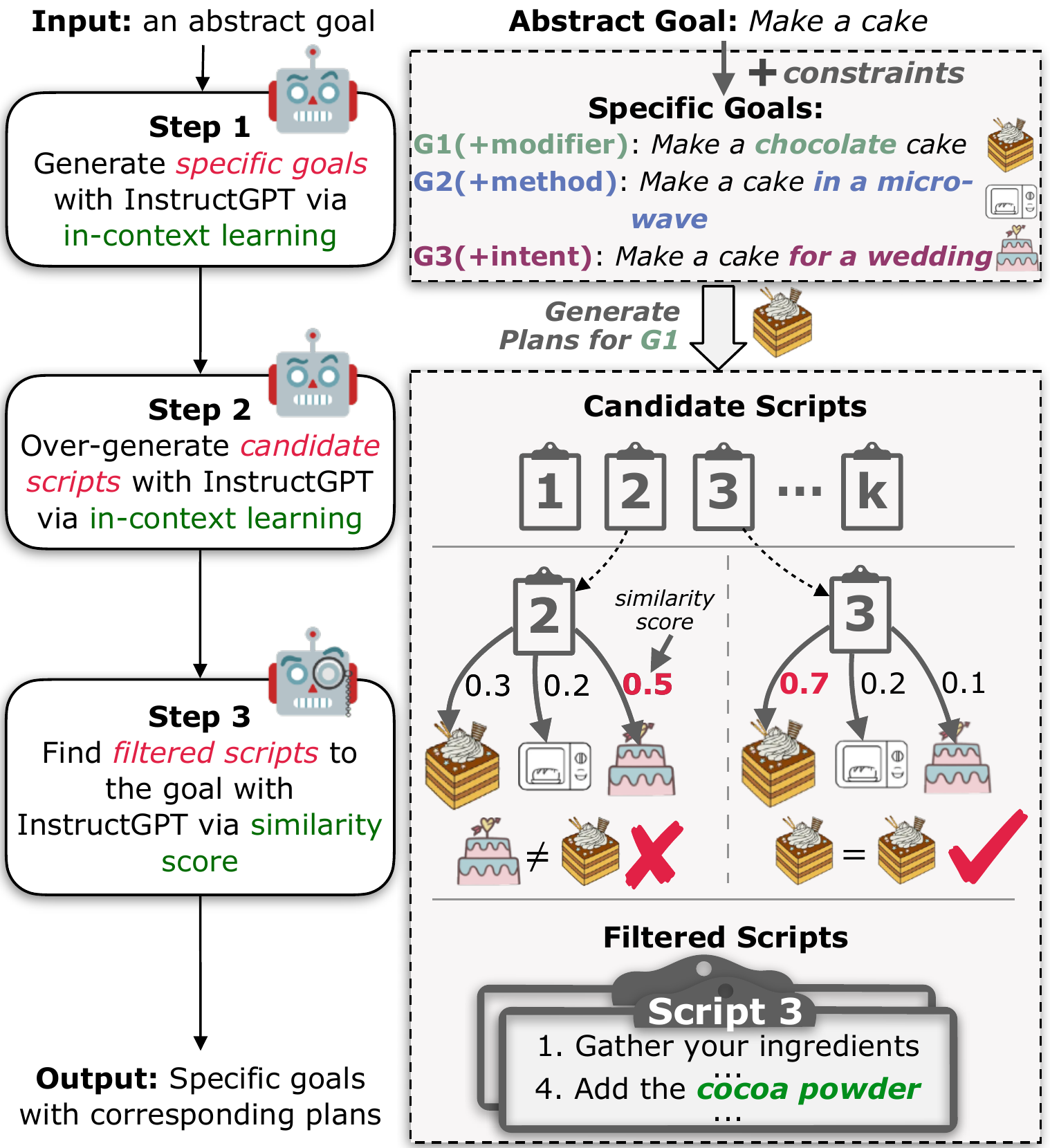}
    \caption{The workflow of using InstructGPT to generate specific goals (Step 1) and planning for the goals with the over-generate-then-filter framework (Step 2-3).}
    \label{fig:pipeline}
\end{figure}

\begin{table}[t]
\small
  \centering
    \begin{tabularx}{\linewidth}{X}
    \toprule
    \rowcolor[gray]{0.95}\multicolumn{1}{c}{\textbf{I: Specific Goal Generation}} \\
    \makecell[l]{
    \color{gray}{/* \textit{Task prompt} */}\\
    Create possible Specific Goals according to the Abstract\\Goal when the Constraint Type is \textit{Modifier}.\\
    \color{gray}{/* \textit{Examples} */} \\
    \textbf{Abstract Goal}: Say Goodbye in Different Language \\
    \textbf{Constraint}: French; \textbf{Specific Goal}: Say Goodbye in French\\
    \textbf{Constraint}: English; \textbf{Specific Goal}: Say Goodbye in English
    \\
    \color{gray}{/* \textit{Auto completion of constraints and specific goals} */} \\
    \textbf{Abstract Goal}: Make a cake \\
    \textbf{Constraint}: \color[rgb]{0,0.39,0}{\textit{Chocolate; Specific Goal: Make a chocolate cake}} \\
    \color[rgb]{0,0.39,0}{\textit{Constraint: Vanilla}}; \textit{Specific Goal: Make a vanilla cake}}\\
    \midrule
    \rowcolor[gray]{0.95}\multicolumn{1}{c}{\textbf{II: Script Generation}} \\
    \makecell[l]{\color{gray}{/* \textit{Task prompt} */}\\
    List the steps of making a cake based on Constraint and\\Specific Goal.\\
    \color{gray}{/* \textit{Examples} */} \\
    \textbf{Goal}: Make a cake \\
    \textbf{Steps}: 1. Gather your ingredients. 2. … \\
    \textbf{Goal}: Make a cupcake \\
    \textbf{Steps}: 1. Decide on a pattern. 2. … \\
    \color{gray}{/* \textit{Auto-completion of script for a specific goal} */} \\
    \textbf{Constraint}: Chocolate;\\ \textbf{Specific Goal}: Make a chocolate cake \\
    \textbf{Steps}: {\color[rgb]{0,0.39,0}\textit{1. Gather ingredients. ... 4. Add the cocoa powder...}}}\\
    \bottomrule
    \end{tabularx}
  \caption{Examples of prompt for InstructGPT for specific goal generation and script generation via in-context learning.
  Generated texts are {\color[rgb]{0,0.39,0}\textit{highlighted}}.
  }
  \label{tab:LLM_prompt}
\end{table}

In this section, we evaluate and enhance the constraint language planning ability of LLMs.
The overall workflow is illustrated in Figure~\ref{fig:pipeline}.
We first extend the specific goals $\mathcal{G}_c$ from the abstract ones $\mathcal{G}_a$ using a human-in-the-loop acquisition approach with LLMs ($\mathsection$~\ref{sec:goal_acquisition}, Step 1), and propose an over-generate-then-filter framework to obtain scripts ($\mathsection$~\ref{sec:our_approach}, Step 2-3).
Then, we reveal that LLMs (\eg, GPT-3~\cite{NEURIPS2020_1457c0d6}, InstructGPT~\cite{ouyang2022training}) are prone to be unfaithful to the constraints in $\mathcal{G}_c$, and our approach can alleviate this problem ($\mathsection$~\ref{sec:exp_llm}).
We use \texttt{text-davinci-002} as the default InstructGPT variant, which has $\ge$175B parameters.\footnote{Code names and approximated parameters of GPT-3 models are based on \url{https://blog.eleuther.ai/gpt3-model-sizes/} and \url{https://beta.openai.com/docs/models}. Note that OpenAI does not release detailed information about later versions of GPT-3, and thus for brevity, we default its size to 175B.}

\subsection{In-Context Learning for LLMs}

\label{sec:Preliminaries}

We deploy LLMs for constrained language planning via in-context learning~\cite{NEURIPS2020_1457c0d6,ouyang2022training}.
Given a task input ($X$), we first write a task prompt ($T$) describing the task, and then provide several examples ($E=\{E_i\}_{i=1}^{|E|}$, where $E_i=(X_i, Y_i)$ are used for few-shot learning).
An LLM generates output ($Y$) by completing the prompt ($Y = \mathcal{M}(T, E, X)$).
The whole process does not require any gradient update, allowing LLMs to generate new specific goals and scripts without massive training data.

\paragraph{Data Source for Examples}
We adopt \wikiHow~\cite{koupaee2018wikihow}, a data source of instructional articles on various topics, as the initial dataset for providing examples.
The articles are titled as ``\textit{how to ...?}'', describing abstract goals, and consist of steps to achieve them.
We use the titles ($\mathcal{G}_a$) and steps ($\mathbf{S}$) as examples.

\subsection{Acquisition of Specific Goals}
\label{sec:goal_acquisition}

Since no dataset of specific goals exists to support our study, we have to acquire these goals first.
As elaborated in Table~\ref{tb:factors}, we extend the abstract goals with multi-faceted constraints for human-in-the-loop data acquisition using InstructGPT.

First, we manually prepare a pool of examples that derive specific goals from an abstract one with constraints.\footnote{Complete examples can be found in Appendix~\ref{appendix:PS_examples}.}
Each example is attached to a constraint type (\ie, modifier, method or intent), and contains more than one constraint and specific goal so that InstructGPT is prompted to generate multiple $\mathcal{G}_c$ for one $\mathcal{G}_a$.
Next, given an abstract goal from \wikiHow, we enumerate each constraint type to ensure data diversity.
Then, we sample several examples of the constraint type from the pool.
Finally, we input the task prompt, examples and the $\mathcal{G}_a$ into InstructGPT for the completion of $\mathcal{G}_c$.

An example in Table~\ref{tab:LLM_prompt} (I) shows InstructGPT generates constraints ``\textit{chocolate}'' and ``\textit{vanilla}'' for $\mathcal{G}_a$ (``\textit{make a cake}'') given the constraint type \textit{modifier} and some examples, and completes the specific goals (``\textit{make a chocolate cake}'' and ``\textit{make a vanilla cake}'').

\subsection{Acquisition of Scripts}
\label{sec:our_approach}

After getting specific goals with constraints, we can test the ability of LLMs to fulfill them.

\paragraph{Planning with InstructGPT}
We first write a task prompt $T$.
Given the $\mathcal{G}_c$, we back-trace its $\mathcal{G}_a$ and extract the verbs (``\textit{make}'') and nouns (``\textit{cake}'') from $\mathcal{G}_a$. 
Then we use the verbs and nouns as keywords to retrieve two similar goals as examples $E$ from the \wikiHow dataset.
Finally, the task prompt $T$, examples $E$ and $\mathcal{G}_c$ with constraint $\mathcal{C}$ are fed into InstructGPT.
As shown in Table~\ref{tab:LLM_prompt} (II), we adopt the scripts, \ie, ``\textit{make a cake}'' and ``\textit{make a cupcake}'' to prompt InstructGPT to generate a script for ``\textit{make a chocolate cake}''.

\paragraph{Over-Generation and Filtering}
Using the above-mentioned approach, generated scripts by InstructGPT are reasonable and fluent.
However, they sometimes are not faithful to the constraints under closer examination ($\mathsection$~\ref{sec:exp_llm}).
Previous studies have shown that the output quality of LLMs falls in high variance~\cite{wiegreffe-etal-2022-reframing}, leading to bad performance.
Thus, we adopt the idea of over-generate-then-filter to improve generation quality, which is shown to be effective in previous work~\cite{wiegreffe-etal-2022-reframing,liu2022wanli}.
We over-generate $K$ sampled from InstructGPT.\footnote{In practice, $K=2$ is sufficient, as shown in Appendix~\ref{appendix:K}. 
Intuitively, the reason this approach works is that 
the generation accuracy can be improved from $1-p$ to $1-p^K$ (at least one is correct), where $p$ is the probability that InstructGPT generates a wrong script.}

Next, a filter model is developed to select the faithful scripts.
Due to the diverse expressions of language, we rely not on rules and patterns (\ie, constraint words must appear in the script), but on the semantic similarity between goals and scripts for filtering.
For example, ``\textit{decorating the cake with candles}'' could be a faithful step to make a cake ``\textit{for a birthday party}''.
Motivated by this, we first collect a set of goals, consisting of the target goal ($\mathcal{G}_{c}^+$) as a positive sample and others ($\{\mathcal{G}_{c}^-\}$) generated from the same abstract goal ($\mathcal{G}_{a}$) as negative samples.
In the previous case, the negatives include ``\textit{make a cake in the microwave}'' and ``\textit{make a cake for a wedding}''.
We convert scripts and goals into InstructGPT embeddings (text-embedding-ada-002) and calculate cosine similarity as similarity scores to measure semantic similarity.  
Additionally, we reward the script that explicitly contains the keywords of the target constraint.
We only keep the script if $\mathcal{G}_{c}^+$ scores the highest in the goal set.

\subsection{Evaluation}
\label{sec:exp_llm}

We randomly collect 100 abstract goals (\eg, ``\textit{make a cake}'') from \wikiHow and conduct manual evaluations on the generated specific goals and their scripts.
We compare our methods with instruction tuning methods, T0~\cite{sanh2022multitask} and Flan-T5~\cite{chung2022scaling}, vanilla GPT-3~\cite{ouyang2022training} with different sizes, Codex~\cite{chen2021evaluating} and InstructGPT~\cite{ouyang2022training} with different sizes.
We also add ``\textit{Let's think step by step}'' before each answer for script generation, which is a simple but effective trick to improve zero-shot reasoning for LLMs~\cite{kojima2022large}.
For a retrieval baseline, we directly use the goals to search and retrieve the most relevant scripts from the \wikiHow website\footnote{https://www.wikihow.com/Main-Page} as results.

\paragraph{Are specific goals generated by LLMs of high quality?}
\label{sec:quality_specific_goals}
We ask InstructGPT to generate 300 ($3\times$) specific goals for 3 constraint types based on the 100 abstract goals from \wikiHow.
For evaluation, we recruit annotators on Amazon Mechanical Turk to check whether these goals are correct.
Each case is examined by three annotators, who reach an inter-rater agreement at Fleiss's $\kappa=0.86$~\cite{fleiss1981measurement}.
InstructGPT achieves 98.00\% accuracy, indicating that LLMs can derive specific goals of rather high quality.

\paragraph{Can LLMs write scripts for specific goals?}
\label{sec:LLMs_goals}
\begin{table}[t]
    \small
    \centering
    \begin{tabular}{lcccc}
    \toprule
        \textbf{Model} & \textbf{Modifier} & \textbf{Method} & \textbf{Intent} & \textbf{All} \\
    \midrule
    Retrieval & 26.67 & 38.89  &  35.71 & 34.00 \\
    \midrule
    T0 (11B) & 30.00 & 21.12 &  25.00 & 24.00 \\
    Flan-T5 (11B) & 50.00 & 42.25 &  31.25 & 42.00 \\
    \midrule
    GPT-3 (1.3B) &13.33 &	12.96 &	18.75 & 14.00 \\
    GPT-3 (6.7B) & 23.33 &  7.40 & 25.00 & 15.00 \\
    GPT-3 (175B) & 30.00 & 22.22 & 25.00 & 25.00 \\
    \cdashlinelr{1-5}
    Codex (175B) & 46.67 & 55.56 & 18.75 & 47.00 \\
    \cdashlinelr{1-5}
    InstructGPT (1.3B) &20.00 & 22.22 & 28.57 & 22.00 \\
    InstructGPT (6.7B) &60.00 & 42.25  & 43.75 & 47.00 \\
    InstructGPT (175B) &73.33 & 74.08  & 42.86 & 69.00 \\
    \ \ + ``\textit{let's think step...}'' &70.00 & 75.92  & 50.00 &68.00 \\
    
    \cdashlinelr{1-5}
     \ \ + Our Method & \textbf{96.67} & \textbf{98.15} & \textbf{92.86} &\textbf{95.00} \\
     \ \ \ \ w/ $f_\mathtt{sim}$ = SBERT &  86.66 & 74.89  &  81.25 & 78.00 \\
     \ \ \ \ w/ $f_\mathtt{sim}$ = SimCSE &  73.33&  78.73 & 75.00 & 75.00 \\
     \ \ \ \ w/ $f_\mathtt{sim}$ = None & 93.33 & 94.44  & 87.50 & 93.00 \\
    \bottomrule
    \end{tabular}
    \caption{Accuracy (\%) of generated scripts for different constraint types by manual evaluation. 
    $f_\mathtt{sim}$ denotes the choice for similarity function during filtering, \ie, replacing InstructGPT embedding with that of SimCSE~\cite{gao-etal-2021-simcse} and Sentence-BERT~\cite{reimers-gurevych-2019-sentence}. \textbf{$f_\mathtt{sim}$ = None} denotes we only reserve the scripts that contain constraint words.}
    \label{tab:LLM_factor}
\end{table}

To answer this question, we first let InstructGPT generate scripts for the 100 abstract goals from \wikiHow and ask three annotators to check the correctness of the scripts (with Fleiss's $\kappa=0.79$).
The correctness is decided by both the fulfillment of the goal and the completeness of the semantics.
InstructGPT achieves 97.00\% accuracy, proving that LLMs can plan for abstract goals very well.
However, it is not the case for specific goals.
We sample 100 specific goals from 300 generated ones (mentioned above) and evaluate the scripts generated from baselines and our method.

Table~\ref{tab:LLM_factor} reports the overall accuracy of the results.
We find that:
\begin{inparaenum}[\it 1)]
    \item Overall, all baselines achieve unsatisfactory results on planning for specific goals, with InstructGPT outperforming others.
    Especially, the scripts with \textit{intent}-type constraints have the worst accuracy, and adding ``\textit{let's think step-by-step}'' does not help much;
    \item The retrieval from \wikiHow does not lead to the desired script;
    \item With our method, InstructGPT can generate scripts of higher quality by a large margin;
    \item Replacing the similarity function with embeddings from other pre-trained models results in performance drops.
\end{inparaenum}

\paragraph{What types of errors do LLMs usually make in this task?}

\begin{figure}[t]
    \centering
    \small
    \includegraphics[width=\linewidth]{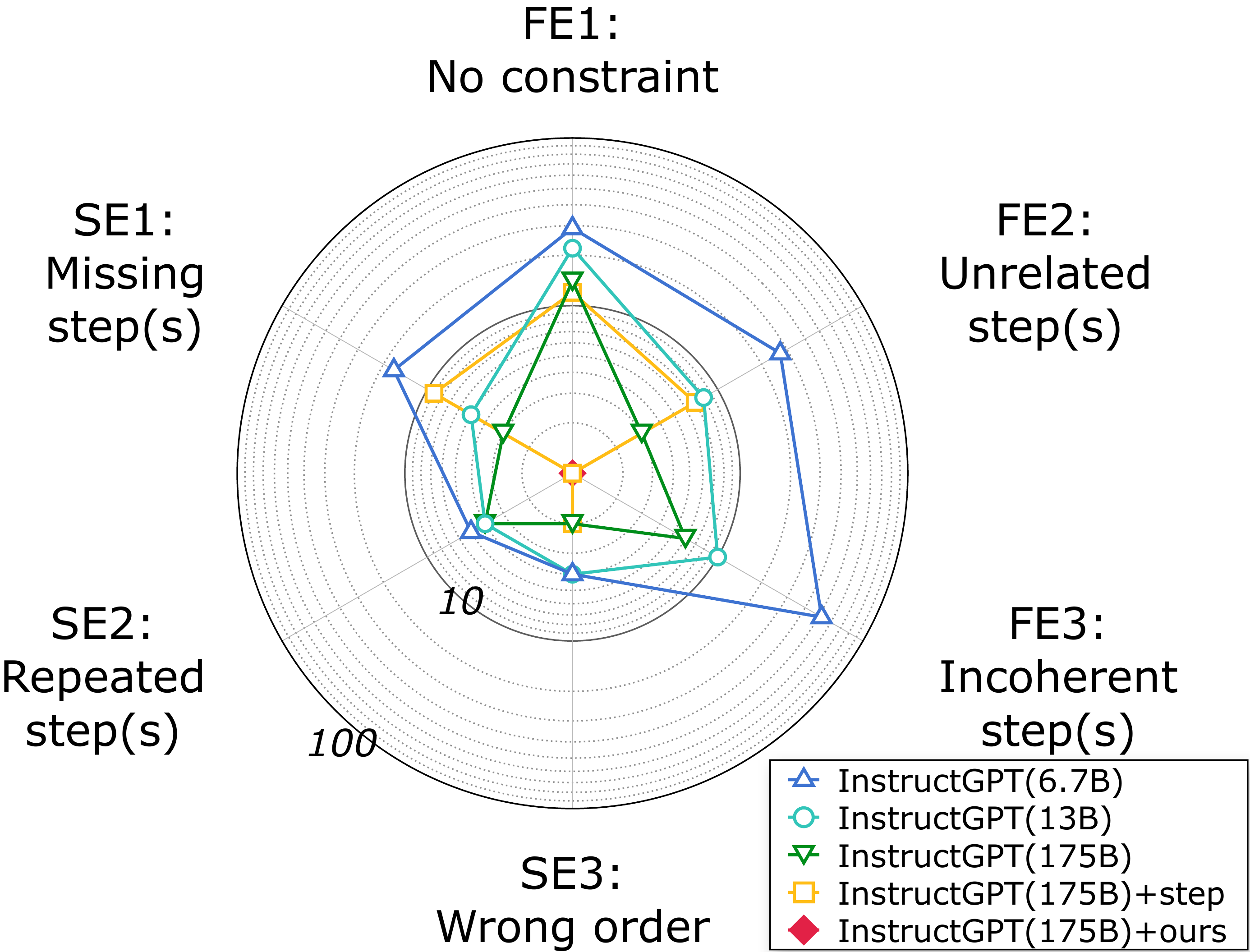}
    \caption{Errors of the generated scripts by human evaluation. The axis of the radar chart is in \textit{log-scale}. Notably, ours reduces to virtually one dot in the graphic because it does not have many errors (0-1\%). SE and FE denote semantic completeness and faithfulness error.}
    \label{fig:LLMs_result}
\end{figure}

To respond to the motivations of our methods, we conduct detailed analyses to investigate why LLMs fail.
We evaluate the model planning performance in two aspects:
\begin{inparaenum}[\it 1)]
    \item \textit{Semantic completeness} (SE): whether the steps in the script are missing, repeated or in the wrong order;
    \item \textit{Faithfulness to the constraints} (FE): whether the script is faithful to the constraints and the steps are coherent (related) within the script.
\end{inparaenum}
We define six types of errors upon the two, \ie, 
\textit{i)} SE: missing, repeated step(s) and wrong order and 
\textit{ii)} FE: no constraint, unrelated step(s) or incoherent step(s).\footnote{The detailed definitions can be found in Appendix~\ref{appendix:type_error}.}
Annotators are asked to review 100 scripts generated by InstructGPT and mark the error types.\footnote{The case study of how InstructGPT fails at planning for specific goals is shown in Appendix~\ref{appendix:case_study}.}
Results in Figure~\ref{fig:LLMs_result} show that:
\begin{inparaenum}[\it 1)]
    \item The semantic completeness in generated scripts is acceptable, but the faithfulness to the constraints can not be guaranteed;
    \item Our method greatly improves the planning quality both in semantic completeness and faithfulness.
\end{inparaenum}

\begin{figure}[t]
    \centering
    \includegraphics[width=\linewidth]{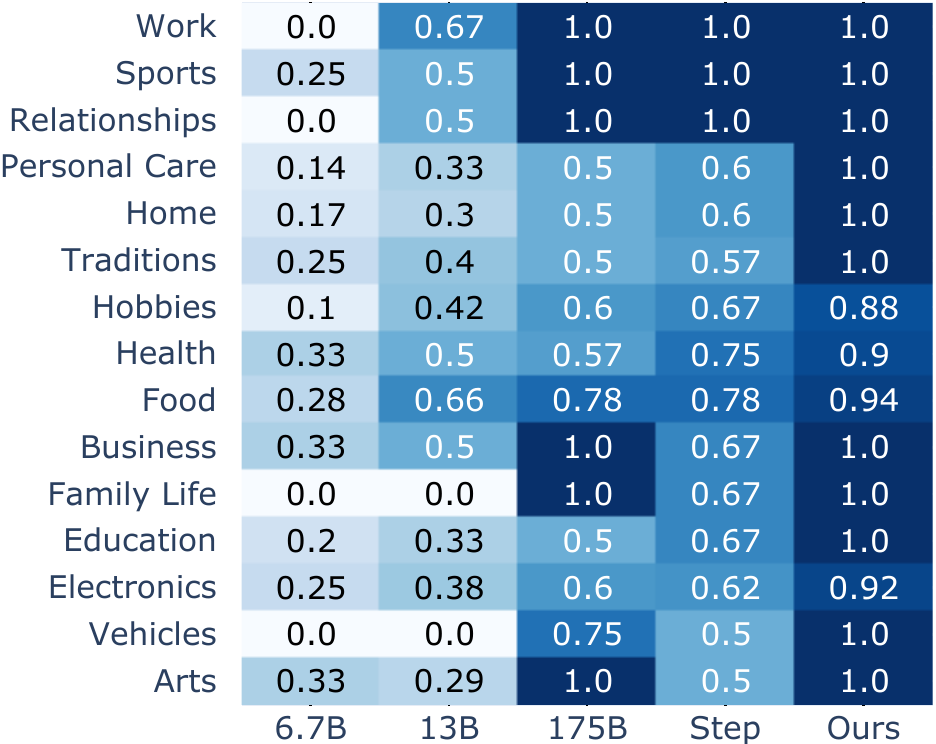}
    \caption{The heat-map depicts the human-evaluated script accuracy of different methods in different topic categories for specific goals.
    }
    \label{fig:LLM_Category}
\end{figure}

\paragraph{What kinds of goals do InstructGPT typically fail?}
By far, we already know that LLMs fail at specific goals, especially for intent-type constraints.
We dig into more fine-grained topic categories of constraints defined in \wikiHow.
The heat map in Figure~\ref{fig:LLM_Category} shows that the planning performance of InstructGPTs varies considerably for goals of different categories, and the planning accuracy for each category improves greatly with our method.

\section{Script Distillation from LLMs}
\label{sec:data}

Since LLMs are costly to deploy, it is essential to enable language planning ability for smaller, specialized models.
Creating datasets is an inevitable step to this end.
However, previous datasets do not enable planning for specific goals~\cite{sakaguchi-etal-2021-proscript-partially,lyu-etal-2021-goal}, and manual dataset annotation is expensive and highly demanding.
Thus, we follow the idea of \textit{symbolic knowledge distillation}~\cite{west-etal-2022-symbolic} to distill constrained language planning datasets from LLMs.

\subsection{\method: A Dataset for Constrained Language Planning}

We now apply our method for building a first-of-its-kind \textbf{Co}nstrained \textbf{Script} dataset of language planning, named as \method.
Experiments in $\mathsection$~\ref{sec:LLM} show that LLMs can generate high-quality specific goals and scripts with our over-generating-then-filter framework.
We now scale up the experiments for a large-scale dataset.
We collect 14,945 article titles as seed abstract goals and retrieve 34,260 similar goals with scripts from \wikiHow as examples to prompt InstructGPT (175B) for data generation.
Following $\mathsection$~\ref{sec:LLM}, dataset construction process consists of three steps, as in Figure~\ref{fig:pipeline}:
1) We first enumerate constraint types with examples for InstructGPT and obtain specific goals (after de-duplication) based on the seed abstract goals.
2) Then, InstructGPT over-generates $K$ scripts for the specific goals and 
3) our filter framework selects the faithful scripts as the final data.\footnote{Details about hyper-parameters and costs can be found in Appendix~\ref{sec:GPT-3_Generation_Hyperparameters}.}

In total, we generate 55,000 specific goals with corresponding scripts.
We randomly choose 2,000 data as the validation set and 3,000 data as the test set.
To ensure the quality of the validation and test set, we ask crowd-sourced workers to find and revise the incorrect samples.
By collecting the annotation data for error identification of these 5,000 samples, we estimate to achieve 97.80\% accuracy for specific goals and 94.98\% for constrained script generation, consistent with the results in Table~\ref{tab:LLM_factor}.

\subsection{Dataset Analysis}


\begin{table}[t]
    \centering
    \small
    \begin{tabular}{lrrrr}
    \toprule
    \textbf{Dataset} & \textbf{\# Size} & \textbf{\# UT} & \textbf{Avg$_{\mathcal{G}_{c}}$ \#} & \textbf{Avg$_{\mathbf{S}}$ \#}\\
    \midrule
        \proScript & 6,414 & 8,826 & 0 & 5.45 \\
        \wikiHow & \textbf{112,111} & \textbf{158,117} & 0.42 & 5.93\\
        \method & 55,000 & 76,317 & \textbf{4.13} &\textbf{5.96} \\
    \bottomrule
    \end{tabular}
    \caption{Statistics of \method and previous script datasets \proScript and \wikiHow, w.r.t. data size, number of unique tokens (\textbf{\# UT}), the average number of specific goals for each abstract ones (\textbf{Avg$_{\mathcal{G}_{c}}$ \#}), and the average number of steps in scripts (\textbf{Avg$_{\mathbf{S}}$ \#}).}
    \label{tab:compare}
\end{table}

\paragraph{Script Diversity Analysis}
As shown in Table~\ref{tab:compare}, despite the larger scale of \wikiHow, \method has more specific goals than \wikiHow and thus is valuable for the constrained language planning task.
Besides, previous studies~\cite{Fu_Lam_So_Shi_2021,narayan-etal-2022-well} find that the texts generated by LMs may be too repetitive and less diverse.
For this concern, we compare our \method with a recent goal-oriented script dataset \proScript~\cite{sakaguchi-etal-2021-proscript-partially} created by crowd-sourcing.
As reported in Table~\ref{tab:compare}, 
\begin{inparaenum}[\it 1)]
    \item \method is much larger than \proScript, with more scripts and a higher number of steps per script;
    \item \method exhibits high lexical diversity, with more unique words than human-written \proScript.
\end{inparaenum}

\begin{figure}[t]
    \centering
    \includegraphics[width=0.7\linewidth]{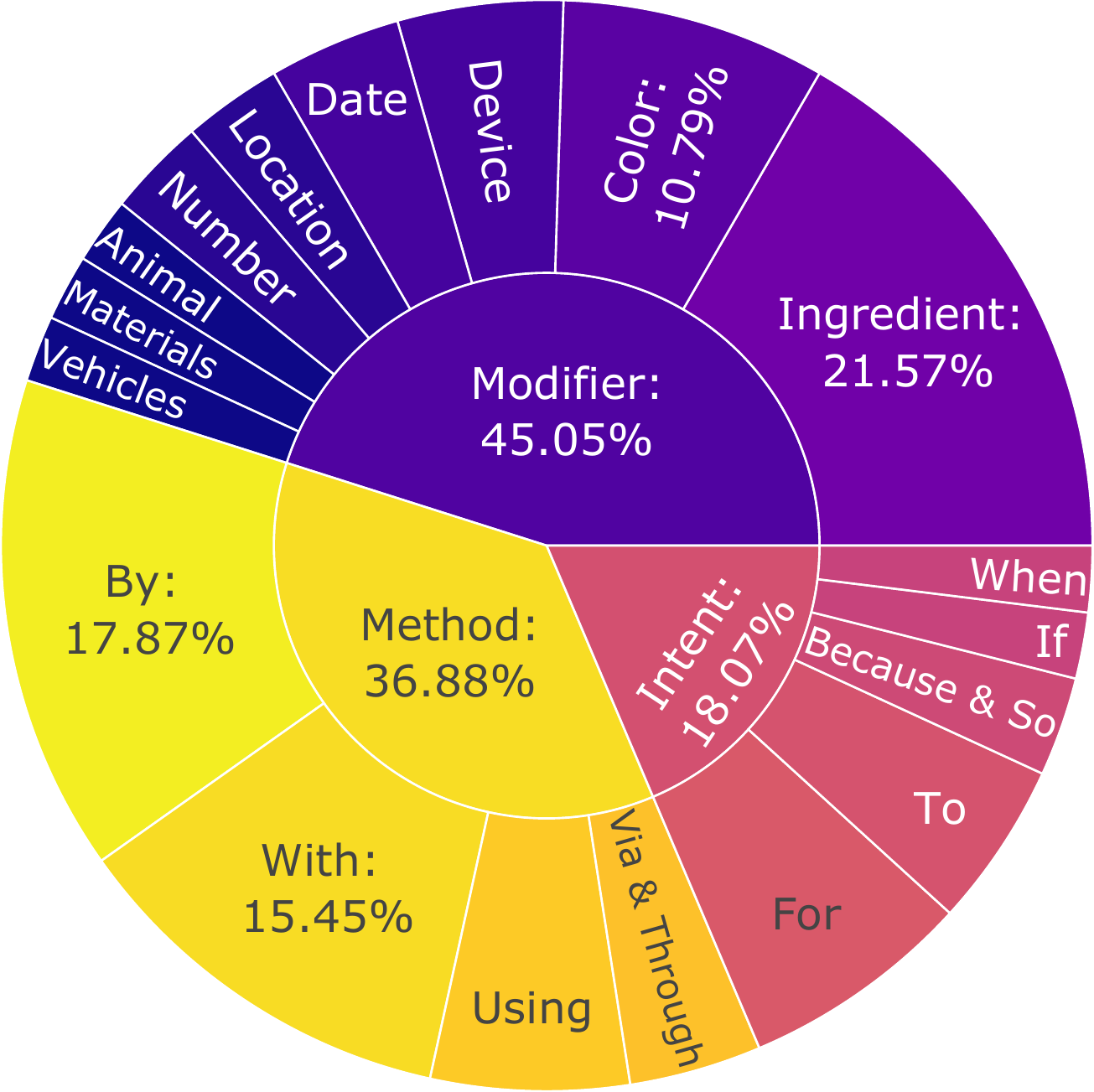}
    \caption{Statistics of constraint types in \method dataset, with representative topic categories or the first words for each constraint type.}
    \label{fig:Distribution}
\end{figure}

\paragraph{Constraint Analysis}
Figure~\ref{fig:Distribution} shows the constraint distribution of \method.
We compute the proportions of constraint types with their representative categories obtained from Probase~\cite{10.1145/2213836.2213891}, and the initial words of constraint instances.
We find \method shows high heterogeneity and pluralism in the generated specific goals.
Interestingly, InstructGPT tends to start with the word ``\textit{if}'' or ``\textit{when}'' for hypothetical constraints (\eg, ``\textit{if someone is lactose intolerant}'' for ``\textit{make a cake}''), suggesting the potential for future research on counterfactual reasoning in language planning.
We also analyze the domain distribution of \method in the Appendix~\ref{appendix:Domain_Examination}

\section{Constrained Language Planning with Specialized Models}
\label{sec:results}

With \method, we can train smaller but specialized models for constrained language planning.


\setlength\tabcolsep{4pt}
\begin{table}[t]
\small
  \centering
    \begin{tabular}{lcccc}
    \toprule
    \textbf{Model} & \textbf{Faithful} & \textbf{ROUGE} & \textbf{BLEU} & \textbf{BERTScore}\\
    \midrule
    \rowcolor[gray]{0.95}\multicolumn{5}{c}{\textit{Trained on \wikiHow}} \\
    GPT-2 & 64.93  & 20.28  & 17.91  & 80.74 \\
    GPT-2 (large)  & 62.20  & 23.74  & 24.69  & 83.63 \\
    \cdashlinelr{1-5}
    T5 (base)  & 86.13  & 20.30  & 15.48  & 79.02 \\
    T5 (large)  & 85.13  & 22.95  & 20.60  & 82.27 \\
    T5 (3B) & 77.90  & 20.72  & 16.95  & 81.01 \\
    \midrule
    \rowcolor[gray]{0.95}\multicolumn{5}{c}{\textit{Trained on \method}} \\
    GPT-2  &  74.60 & 28.09  & 26.75  & 84.72 \\
    GPT-2 (large)  & \textbf{76.73} & 30.60  & 30.22  & 85.77 \\
    \ \ +\textit{retrieval} & 76.30  & \textbf{32.78 } & \textbf{32.92 } & \textbf{86.41} \\
    \cdashlinelr{1-5}
    T5 (base)  & 91.53 & 26.53  & 22.06  & 83.14 \\
    T5 (large)  & 91.87  & 29.40  & 29.14  & 83.48 \\
    \ \ +\textit{retrieval} &  86.03 & 35.91 & 36.10 & 87.39 \\
    T5 (3B) & \textbf{93.00} & 45.68  & 43.83 & 90.18 \\
    \ \ +\textit{retrieval} & 92.53 & \textbf{46.54}  & \textbf{47.62} & \textbf{90.84} \\
    \bottomrule
    \end{tabular}
  \caption{Overall script generation performance for models trained on different training sets. Note that the test set is the same for all models.}
  \label{tab:SLM}
\end{table}

\subsection{Experimental setup}

\paragraph{Baselines}
We use GPT-2 (causal LM)~\cite{radford2019language} and T5 (encoder-decoder LM)~\cite{raffel2020exploring} as baselines.
Given goals, the models are trained to generate a list of steps $\mathbf{S}$ for planning.
Moreover, we adopt the idea of retrieval-augmented text generation~\cite{NEURIPS2020_6b493230} and add retrieved examples in the input to improve generation quality.


\paragraph{Metrics}
We use BLEU~\cite{papineni-etal-2002-bleu}, ROUGE-L~\cite{lin-2004-rouge} and BERTScore~\cite{Zhang*2020BERTScore:} as automatic metrics to measure semantic completeness.
We also train a binary classification model to decide whether the generated texts are faithful to the constraints. 
Specifically, we collect 50,000 data from \method as positive samples, and shuffle the goals and scripts to construct 50,000 negative ones. 
Then, we fine-tune a DeBERTa (v3 large) model~\cite{khashabi-etal-2020-unifiedqa} for classification, achieving 91.53\% accuracy on the test set.

\paragraph{Training Data}\label{sec:training_data}
To gain a fine-grained perspective on planning toward specific goals, we train LMs on both \wikiHow ($\mathbb{D}_\mathtt{tr}^\texttt{wi}$) and \method ($\mathbb{D}_\mathtt{tr}^\texttt{co}$), and test them on \method test set ($\mathbb{D}_\mathtt{te}^\texttt{co}$).
Both datasets share \textit{similar} scripts, but the goals in \wikiHow are mostly abstract ones.
For \wikiHow, we also randomly collect 50,000 goals with scripts as $\mathbb{D}_\mathtt{tr}^\texttt{wi}$.

\subsection{Results}\label{sec:result_SLM}

\begin{table}[t]
    \small
    \centering
    \begin{tabular}{lcccc}
    \toprule
        \textbf{Model} & \textbf{Modifier} & \textbf{Method} & \textbf{Intent} & \textbf{All} \\
    \midrule
    T5 (large)  & \textbf{91.54} & \textbf{92.57} & \textbf{90.21} & \textbf{91.81}\\
    \ \ +\textit{retrieval} & 87.39 & 85.86 & 84.44 & 86.03\\
    \cdashlinelr{1-5}
    GPT-2 (large)  & \textbf{78.78} & \textbf{78.77} & 69.48 & \textbf{76.73}\\
    \ \ +\textit{retrieval} & 77.33 & 78.28 & \textbf{70.97} & 76.30\\
    \bottomrule
    \end{tabular}
    \caption{\textbf{Faithfulness} scores of specialized models for each constraint type on the test set of \method.}
    \label{tab:SLM_factor}
\end{table}

The comparison for models trained on \wikiHow and \method are shown in Table~\ref{tab:SLM}.
In general, LMs trained on \method outperform that on \wikiHow.
T5 outperforms GPT-2 in faithfulness, possibly due to its encoder-decoder framework being better at handling input information. 
However, GPT-2 outperforms T5 on other text generation metrics for scripts.
This could be because \method is distilled from InstructGPT, leading to a biased data distribution that favors decoder-only causal language models, \eg, the GPT family.

Based on Table~\ref{tab:SLM}, we find that augmenting models with retrieved examples can improve semantic completeness.
However, the constraint faithfulness could be undermined as models tend to mimic the retrieved examples.
To further understand the role of retrieval augmentation, we conduct a manual evaluation that based on 100 random samples generated by T5 (3B) with and without retrieval augmentation. 
We discover that 57\% of T5's results are correct, and the number goes up to 70\% with retrieval augmentation.
Thus, although we observe a slight drop in faithfulness score ($93.00\rightarrow 92.53$ from Table~\ref{tab:SLM}), retrieval augmentation still brings much improvement over the base model.

\paragraph{Faithfulness of Constraints of Different Types}
Will LLMs' planning preferences for constraint types pass to the specialized models?
We find the results in Table~\ref{tab:SLM_factor} are consistent with that of LLMs (Table~\ref{tab:LLM_factor}).
Specialized models are also the worst at specific goals with intent-typed constraints.

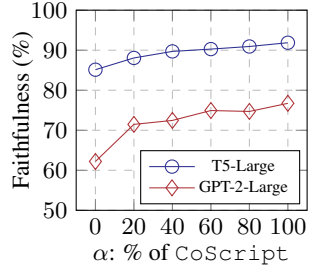
\begin{figure}[t]
    \centering
\pgfplotsset{width=0.6\linewidth,height=0.55\linewidth,compat=1.15}
\footnotesize
\begin{tikzpicture}
\begin{axis}[
    xlabel={$\alpha$: \% of \method},
    ylabel={Faithfulness (\%)},
    xmin=-10, xmax=110,
    ymin=50, ymax=100,
    xtick={0, 20, 40, 60, 80, 100},
    ytick={50, 60, 70, 80, 90, 100},
    legend pos=south east,
    ymajorgrids=true,
    xmajorgrids=true,
    grid style=dashed,
    x label style={at={(axis description cs:0.5,-0.125)},anchor=north},
    y label style={at={(axis description cs:-0.15,0.5)},anchor=south},
    legend style={nodes={scale=0.7, transform shape}}
]
\addplot[
    color=Blue,
    mark=o,
    mark size=2.5pt,
    ]
    coordinates {
    (0, 85.13)
    (20, 88.07)
    (40, 89.70)
    (60, 90.28)
    (80, 90.93)
    (100, 91.87)
    };
    \addlegendentry{T5-Large}
    
\addplot[
    color=Maroon,
    mark=diamond,
    mark size=3pt,
    ]
    coordinates {
    (0, 62.20)
    (20, 71.47)
    (40, 72.46)
    (60, 74.93)
    (80, 74.70)
    (100, 76.73)
    };
    \addlegendentry{GPT-2-Large}

\end{axis}
\end{tikzpicture}

    \caption{The faithfulness curves when altering the proportions of \method ($\alpha$) and \wikiHow ($1-\alpha$) in a fixed-size training set.
    }
    \label{fig:Rheostat}
\end{figure}

\setlength\tabcolsep{2pt}
\begin{table}[t]
    \small
    \centering
    \begin{tabular}{lccccc}
    \toprule
        \textbf{Model} & \textbf{Size} & \textbf{Modifier} & \textbf{Method} & \textbf{Intent} & \textbf{All} \\
    \midrule
    GPT-3 & 175B & 30.00 & 22.22 & 25.00 & 25.00 \\
    Codex & 175B & 46.67 & 55.56 & 18.75 & 47.00 \\
    InstructGPT & 175B & {73.33} & \textbf{74.08}  & 42.86 & {69.00} \\
    \cdashlinelr{1-6}
    T5 (\wikiHow) & 3B  & 20.00  & 12.96 & 6.25 & 14.00 \\
    T5 (\method) & 3B & 63.33 & 55.55 & {43.75} & 56.00 \\
    \ \ +\textit{retrieval} & 3B & \textbf{76.66} & {66.66} & \textbf{75.00} & \textbf{71.00} \\
    \bottomrule
    \end{tabular}
    \caption{
   \textbf{Accuracy} (\%) of scripts generated by different models.
   We fine-tune a T5 (3B) on \wikiHow and \method while deploying LLMs via few-shot in-context learning.
    }
    \label{tab:T5_factor}
\end{table}
\paragraph{\method vs. \wikiHow}

We mix two datasets together with a hyper-parameter $\alpha$ to control the proportion of two datasets, where the new training set $\mathbb{D}_\mathtt{tr} = \alpha \mathbb{D}_\mathtt{tr}^\texttt{co} + (1-\alpha)\mathbb{D}_\mathtt{tr}^\texttt{wi}$.
By altering $\alpha$ (constant data size), the faithfulness curves in Figure~\ref{fig:Rheostat} shows that adding more data from \method consistently improves model performance in constraint faithfulness.
Thus, training on \method contributes to more faithful planners.

\paragraph{Specialized Models vs. LLMs}

We further fine-tune a T5 (3B) on \method and \wikiHow to generate scripts for the specific goals in $\mathsection$~\ref{sec:exp_llm}, which are held out from the training set.
Table~\ref{tab:T5_factor} shows that T5 fine-tuned on \method with retrieval augmentation can generate scripts of higher quality than most LLMs in Table~\ref{tab:LLM_factor}, indicating that smaller models can surpass larger models when properly trained on suitable datasets.

\section{Conclusion}
\label{sec:conclusion}

In this paper, we define planning toward specific goals with constraints.
We propose a better prompting method for LLMs, and distill a novel dataset from LLMs (\method) to improve the constrained language planning ability of specialized models.
Experiments show that our method improves the planning quality of LLMs for specific goals, and smaller models trained on \method even outperform LLMs.
We hope the \method dataset will be a valuable resource to advance the research on language planning with more complex and diverse goals and constraints.


\section*{Limitations}
\label{sec:limitation}

The proposed method for improving LLMs is a post-hoc re-ranking approach, and we do not improve LLMs themselves due to the difficulty of fine-tuning LLMs.
Besides, we improve the ability of constrained language planning for smaller models from the perspective of building task-related datasets, but do not consider investigating the model itself, other than adopting retrieval augmentation.
In addition, because automatic metrics for generated text are limited, the automatic evaluation of this paper may result in an overestimation or underestimation of the mentioned methods, though we attempt to mitigate this by incorporating a moderate amount of human evaluation.
Despite the advanced planning capabilities of newer language models, our work remains significantly valuable to the knowledge distillation of these LLMs into smaller and more cost-effective models.

We also discover several limitations of the proposed \method datasets.
First, the specific goal explored in this work only inherits from an abstract one with one extra constraint.
However, in real-life situations, complex planning may involve multiple constraints, which we do not investigate in this work.
Another limitation of \method is that our dataset is generated from InstructGPT, and thus the data distributions may be biased to favor causal language models.
This is a common issue with machine-generated datasets, which we address by manually curating \method's validation and test sets.
Furthermore, there are still some incorrect samples (about 5\%) in the training data without manual correction due to the limits of budget and time.
Last but not least, we only consider whether the script can be executed at the human level. 
The script execution for robots~\cite{huang2022language,lu2022neuro} is unstudied in our work, and there still exist huge gaps in transferring complex human language to one that is understandable and executable by robots.

\section*{Ethics Statement}
\label{sec:Ethics}
\paragraph{Use of Human Annotations}
We protect the privacy rights of crowd-sourced workers and pay them above the local minimum wage.
We use Amazon Mechanical Turk (AMT) and require 300 annotators to be located in the U.S. as a proxy for English competency. We pay at a rate of \$6/hour for 20 samples. 
We acknowledge that constructing datasets from large language models may suffer from toxic language and cause severe risks for social society~\cite{ousidhoum-etal-2021-probing,baldini-etal-2022-fairness}.
Therefore, we ask the annotators to discard the offensive and harmful data when reviewing the \method.
However, there may still be prejudicial data in our final dataset that goes unnoticed.

\paragraph{\wikiHow Source}
The content available on wikiHow is shared under a Creative Commons License (CC-BY-NC-SA)~\footnote{https://creativecommons.org/licenses/by-nc-sa/3.0/}, which permits others to share, copy, distribute, and adapt the content for non-commercial purposes.
In our research, we use \wikiHow as an initial dataset for providing examples to construct our dataset.
Our dataset is released on GitHub and is only used to advance academic research on language planning with more complex and diverse goals and constraints.
Therefore, we emphasize that our usage aligns with the requirements under the license.

\paragraph{Covered Domains in \method}
\method is derived from wikiHow and encompasses 19 daily life goal categories (as illustrated in Figure~\ref{fig:domain_category}). 
These categories cover a wide range of practical topics of everyday life. 
However, as shown in Figure~\ref{fig:domain_category}, we emphasize that sensitive and high-risk domains, including medical, legal, and high-stakes financial advice, are excluded from the dataset to minimize potential risks related to inaccurate or misleading information. 
We encourage researchers and developers to leverage this dataset to build models that accurately understand and respond to user queries on various non-sensitive, non-critical topics.

\paragraph{Factuality, Toxicity and Biases}
We recognize that the factuality of generated content is crucial, especially in high-stakes scenarios. 
Therefore, annotators are asked to verify the consistency between generated scripts and goals with constraints for validation and test sets.
They also assess and revise the content to minimize hallucinations, factual errors, and any inappropriate or misleading information.

Previous work found that LLMs may generate toxic contents~\cite{cao-etal-2022-hallucinated,liu-etal-2022-token}.
We highlight that our dataset is not intended for safety-critical applications or as a substitute for expert advice in such domains. 
Annotators are specifically instructed to discard offensive and harmful data during the review of the validation and test sets in \method. 
However, despite these precautions, there may still be some prejudicial data that goes unnoticed in our final dataset.

\section*{Acknowledgement}
We thank the anonymous reviewers for their valuable comments, and Wei Shi and Shuang Li from Fudan University for their useful suggestions for the manuscript.
This work is supported by the Chinese NSF Major Research Plan (No.92270121), Shanghai Science and Technology Innovation Action Plan (No.21511100401) and the Science and Technology Commission of Shanghai Municipality Grant (No. 22511105902).

\bibliography{anthology}

\appendix

\section{Author Contributions}

\paragraph{Siyu Yuan}
Lead the project,
develop the original method and original code,
lead the curation of the dataset,
contribute to the original experiments, and
contribute to the original manuscript.

\paragraph{Jiangjie Chen}
Conceptualization of the original idea, 
supervision of the research activity planning and execution,
contribution to the original manuscript and figures, and
acquisition of financial support for the project.

\paragraph{Ziquan Fu}
Contribute to the original idea,
provide financial support for the project,
revise the manuscript, and
contribute to data curation.

\paragraph{Xuyang Ge}
Contribute to the experiments and data curation.

\paragraph{Soham Shah}
Provide financial support for the project and
revise the manuscript.

\paragraph{Charles Robert Jankowski}
Provide financial support for the project and
revise the manuscript.

\paragraph{Yanghua Xiao} 
Provide financial support for the project and
revise the manuscript.

\paragraph{Deqing Yang}
Provide financial support for the project,
revise the manuscript, and
oversee the research activity execution.


\section{Implementation Details}

\subsection{Handpicked Examples for Specific Goal Generation}\label{appendix:PS_examples}

We follow the instructions proposed by \citet{mishra-etal-2022-reframing} to better construct the three examples for specific goal generation.
As shown in Table~\ref{tab:handpicked}, we turn long descriptions into bulleted lists in the task prompt for better generation.
In addition, in each example, we list two specific goals with constraints, which can prompt InstructGPT to generate multiple specific goals for the given abstract goals.
In our experiment, we conduct the specific goals generation with different numbers of examples and report the accuracy and the total number of generated specific goals for 100 abstract goals. 
The results in Table~\ref{table:num_example_PS} show that three examples are good in our settings.

\begin{table}[!htb]
  \centering
  \small
    \begin{tabular}{ccc}
    \toprule
    \textbf{\# Examples} & \textbf{Accuracy} & \textbf{\# Total} \\
    \midrule
    2     & 95.16\% & \textbf{545} \\
    3     & \textbf{96.99\%} & 537 \\
    4     & 95.44\% & 539 \\
    \bottomrule
    \end{tabular}%
    \caption{The specific goals generation performance of InstructGPT with different numbers of examples.}
    \label{table:num_example_PS}
\end{table}%

\subsection{Prompts Format}

To explore the prompt formats on script generation, we test 100 samples mentioned in $\mathsection$~\ref{sec:exp_llm} without using the task prompt or replacing the original words with other words. 
The results in Table~\ref{tab:prompt_format} show that:
\begin{inparaenum}[\it 1)]
    \item task prompt can help InstructGPT to better understand the task and thus can improve the model performance on script generation;
    \item adopting popular words to write the prompts can better improve the effect of prompts.
\end{inparaenum}

\begin{table}[!htb]
    \small
    \centering
    \begin{tabular}{lccc}
    \toprule
        \textbf{Format} & \textbf{Goal} & \textbf{Script} \\
    \midrule
    Our Method &  98.00 &  69.00\\
    \cdashlinelr{1-4}
    \ \ \ \ \wo \textit{Task Prompt} &  94.67 &  64.00\\
    \cdashlinelr{1-4}
    \ \ \ \ \textit{r.} \textit{Goal$\rightarrow$Scenario} &  96.00 &  -\\
    \ \ \ \ \textit{r.} \textit{Abstract$\rightarrow$General} &  96.67 &  -\\
     \ \ \ \ \textit{r.} \textit{Step$\rightarrow$Event} &  - &  67.00\\
    \bottomrule
    \end{tabular}
    \caption{Accuracy (\%) of different prompt formats by manual evaluation. We replace (\textbf{r.}) the words in Table~\ref{tab:LLM_prompt} with other words for comparison.}
    \label{tab:prompt_format}
\end{table}

\subsection{Over-generation Hyper-parameters}\label{appendix:K}

\begin{figure}[!htb]
    \centering
    \includegraphics[width=0.85\linewidth]{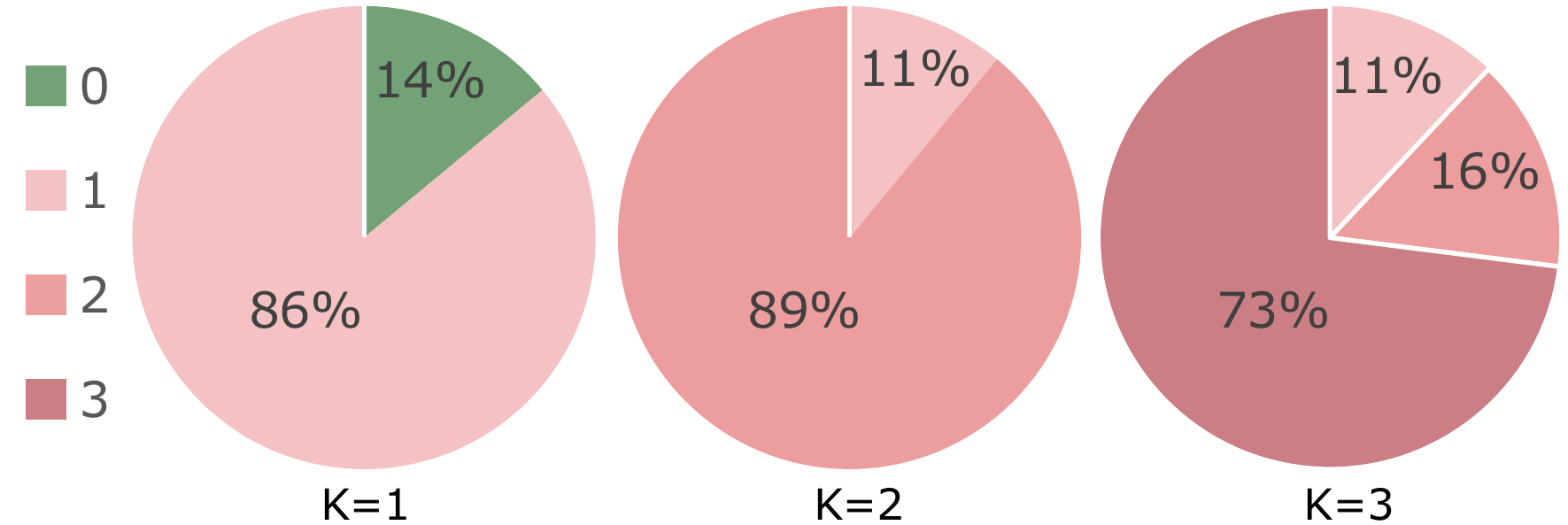}
    \caption{Statistics of the constraints evaluation results. 
    Human annotators are asked to evaluate $K$ generations for each specific goal on 100 samples. 
    The legends indicate the number of constraint-faithful scripts in K over-generation.
    We report the proportion of generated scripts faithful to the constraints.}
    \label{fig:pie}
\end{figure}

To evaluate the LLMs on planning for specific goals, we randomly sample 100 specific goals and then generate scripts from the baseline and our method.
Figure~\ref{fig:pie} reports the faithfulness to constraints.
We find that:
\begin{inparaenum}[\it 1)]
    \item the output quality of InstructGPT falls in high variance, and the script may be unfaithful to the constraints;
    \item over-generation can amplify the likelihood of constraint satisfaction, and $K=2$ is sufficient.
\end{inparaenum}

\subsection{Error Types}\label{appendix:type_error}
As shown in Figure~\ref{tab:type_error}, we evaluate the model planning performance on two aspects, \ie, \textit{Semantic completeness} and \textit{Faithfulness to the constraints} ($\mathsection$~\ref{sec:exp_llm}), and define six types of errors. 

\begin{table*}[t]
    \small
    \centering
    \begin{tabular}{cllll}
    \toprule
    \textbf{Aspects}& \textbf{Error Types} & \textbf{Explanation} & \multicolumn{2}{c}{\makecell{\textbf{Example: Make a vanilla cake}\\\textbf{Constraint}: \colorbox[rgb]{0.99, 0.95, 0.93}{\color[rgb]{0.81, 0.41, 0.22}Vanilla}}} \\
    \midrule
    \multicolumn{1}{c}{\multirow{3}[6]{*}{\makecell{Semantic\\Completeness}}} & {\color{violet}Wrong order} & \makecell[l]{Steps that are in \\the wrong order.} & \multicolumn{1}{l}{\multirow{6}{*}{\makecell[l]{Correct Script:\\1. Gather your ingredients.\\2. Preheat the oven to 325° F \\and grease and flour a cake pan.\\3. Cream the butter and suger.\\4. Add the eggs and vanilla.\\5. Stir in the cake flour.\\6. Pour the batter into the pan.\\7. Bake the cake for 1 hour \\ \ \ \ 15 minutes.}}} & \multicolumn{1}{l}{\multirow{6}{*}{\makecell[l]{Generated Script:\\{\color{violet}1. Preheat the oven to 325° F} \\{\color{violet}and grease and flour a cake pan.}\\{\color{violet}2. Gather your ingredients.}\\3.   {\color{orange}Buy your ingredients.}\\{\color{red}4. Cream the butter and }{\color{teal}salt.}\\{\color{red}5. Stir in the cake flour.}\\6.   {\color{blue}Have a shower.}\\7.   Pour the batter into the pan.\\8.   Bake the cake for 1 hour\\ \ \ \ 15 minutes.}}} \\
\cmidrule{2-3}          & {\color{orange}Repeat steps} & \makecell[l]{Steps that are \\repeated in the script.} &       &  \\
\cmidrule{2-3}          & {\color{red}Missing steps} & \makecell[l]{Important steps that\\ are missing.} &       &  \\
\cmidrule{1-3}    \multicolumn{1}{c}{\multirow{3}[6]{*}{\makecell{Faithfulness\\to Constraints}}} & \colorbox[rgb]{0.99, 0.95, 0.93}{\color[rgb]{0.81, 0.41, 0.22}No constraint} & \makecell[l]{Script is unfaithful\\ to the constraint} &       &  \\
\cmidrule{2-3}          & {\color{teal}Incoherent steps} & \makecell[l]{Steps that are related\\ to the goal, but are not\\ coherent within the script.} &       &  \\
\cmidrule{2-3}          & {\color{blue}Unrelated steps} & \makecell[l]{Steps that are not\\ related to the goal.} &       &  \\
    \bottomrule
    \end{tabular}%
    \caption{The error types and their explanation with examples.}
    \label{tab:type_error}
\end{table*}

\subsection{Case Study}\label{appendix:case_study}
Table~\ref{tb:case_study} lists three examples by InstructGPT (175B) and our approach. 
The first and second examples show that the scripts generated by InstructGPT may fail in unfaithfulness to the constraints.
The third examples demonstrate that although the scripts generated by InstructGPT can be faithful to the constraints, they may suffer from other error types.
In contrast, the over-generating-and-filtering method can amplify the likelihood of high-quality generations and thus can make LLMs better planners.

\section{\method Details}

\subsection{Generation Hyper-parameters}\label{sec:GPT-3_Generation_Hyperparameters}
We queried the \texttt{text-davinci-002} model through the OpenAI API on June 25 to July 5, 2022. 
\method is generated under a specified license that is compatible with the conditions under OpenAI API.
In total, the generation for \method costs about \$5,000. 
The hyper-parameters for script generation are shown in Table~\ref{table:hyperparameters}.
If both generations are faithful, we randomly select one into the dataset.

\begin{table}[!htb]
    \centering
    \small
\begin{tabular}{lc}
\toprule
\textbf{Hyper-parameter} & \textbf{Assignment} \\
\midrule
Top-$p$                   & 1.0                   \\
Temperature             & 1.0                   \\
Max tokens              & 512                 \\
Presence penalty        & 0.0                 \\
Frequency penalty       & 0.0                 \\
$K$                       & 2                   \\ 
\bottomrule
\end{tabular}
\caption{Hyper-parameters for script generation from InstructGPT.}
    \label{table:hyperparameters}
\end{table}

\begin{figure}[t]
    \centering
    \includegraphics[width=\linewidth]{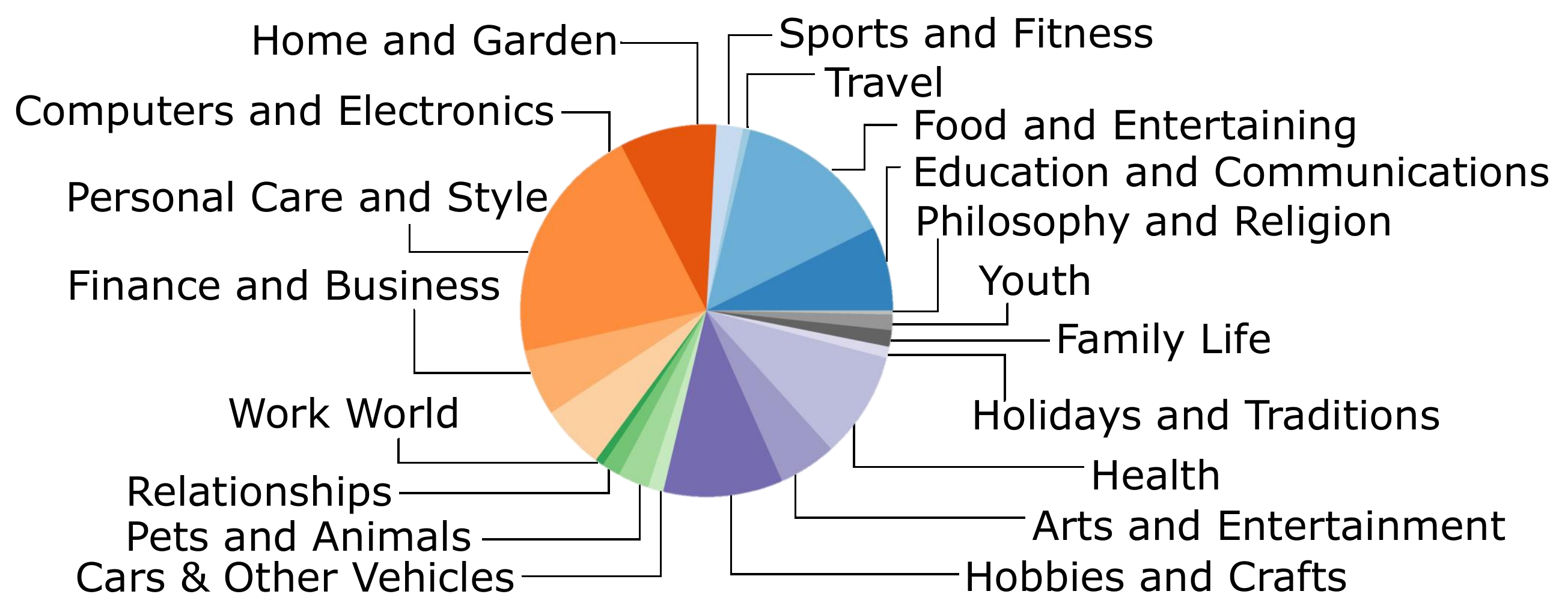}
    \caption{The category distribution of \method. The categories are derived from \wikiHow}
    \label{fig:domain_category}
\end{figure}

\begin{table}[t]
\footnotesize
  \centering
    \begin{tabularx}{\linewidth}{X}
    \toprule
    \makecell[l]{
    \color{gray}{/* \textit{Task prompt} */}\\
    Given a goal, please carefully evaluate and determine if\\ it falls under sensitive and high-risk domains, specifically \\within the fields of medical and legal professions.\\
    \color{gray}{/* \textit{Examples} */} \\
    Goal: Treat COVID-19 at Home\\
    Answer: Yes\\
    Goal: Make Strawberry Cake with Microwave \\
    Answer: No \\
    \color{gray}{/* \textit{Test data} */} \\
    Goal: Clean Oil Stains from Clothes with Soap\\ 
    Answer: \color[rgb]{0,0.39,0}\textbf{\textit{No}}
    } \\
    \bottomrule
    \end{tabularx}
  \caption{The instruction for ChatGPT to identify sensitive and high-risk domains. Generated texts by ChatGPT are {\color[rgb]{0,0.39,0}\textit{\textbf{highlighted}}}.
  }
  \label{tab:chatgpt}
\end{table}
\subsection{Domain Examination}\label{appendix:Domain_Examination}
As shown in Figure~\ref{fig:domain_category}, \method is derived from \wikiHow and encompasses 19 daily life goal categories. 
These categories cover a wide range of practical topics of everyday life, excluding sensitive and high-stakes topics like medical and legal matters.
In addition, we adopt ChatGPT to assess each specific goal in \method to further mitigate risks associated with sensitive domains.
The instruction is shown in Table~\ref{tab:chatgpt}.
Upon examination, ChatGPT identifies that 0.24\% of specific goals (132) within the \method involve a sensitive domain, such as "relieve head pain with medication" in the \textit{Health} domain. 
We manually remove these data and substitute them with new specific goals (\eg, relieve head pain with meditation) to ensure the safety of our dataset.
We encourage researchers and developers to leverage this dataset to build models that accurately understand and respond to user queries on non-sensitive, non-critical topics.

\subsection{Qualitative Generations}
We randomly select qualitative generations from the \method. Table~\ref{table:PS_example_modifier}, Table~\ref{table:PS_example_method} and Table~\ref{table:PS_example_intent} show some specific goal generations under different types of constraints. 
Table~\ref{table:GE_example} shows some scripts generated based on our proposed pipeline. 

\section{Crowd-sourcing Details}\label{appendix:crowdsourcing}
\begin{figure*}[t]
    \centering
    \includegraphics[width=\linewidth]{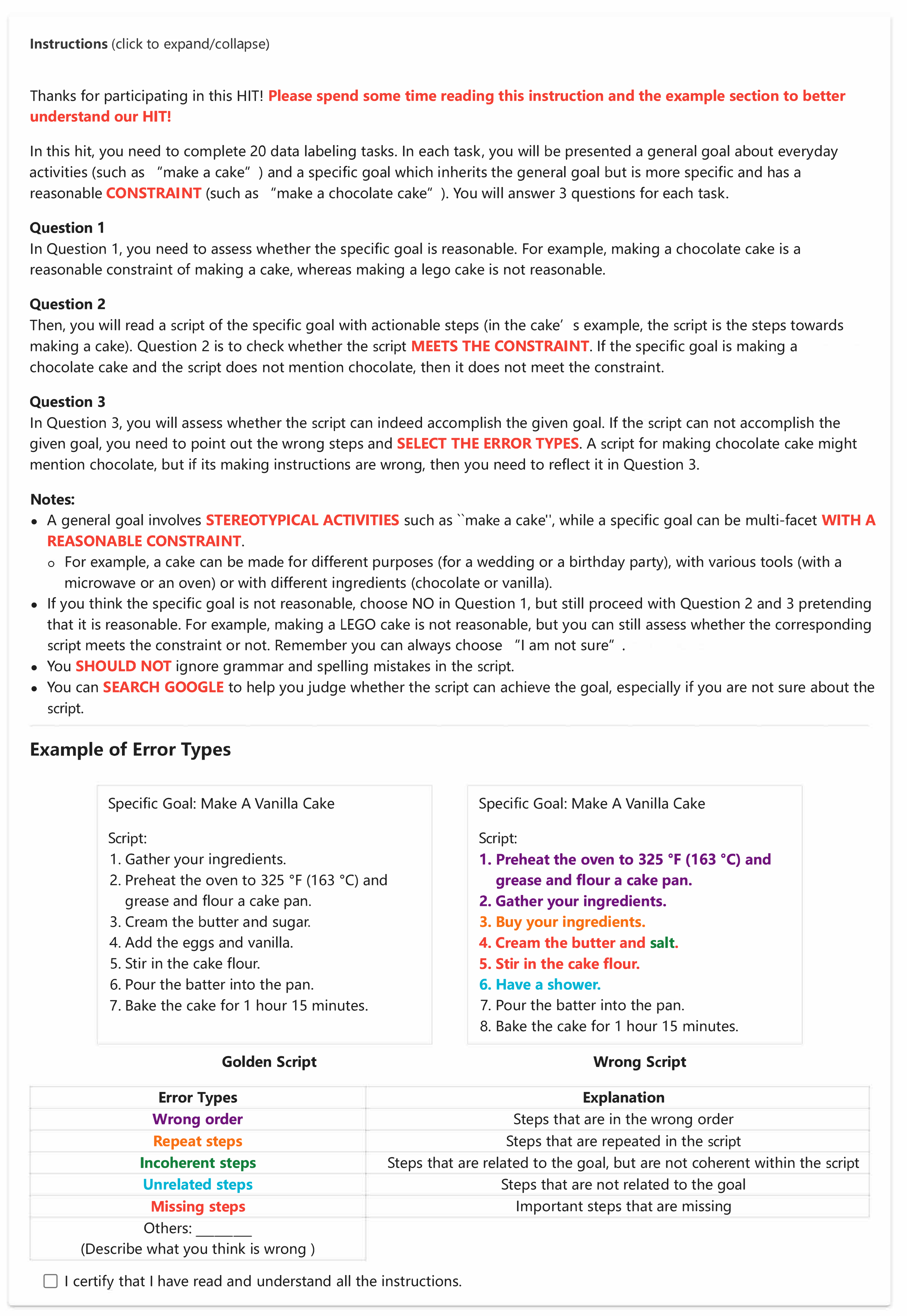}
    \caption{Instructions for crowd workers on Amazon Mechanical Turk.}
    \label{fig:instructions}
\end{figure*}

\begin{figure*}[t]
    \centering
    \includegraphics[width=0.85\linewidth]{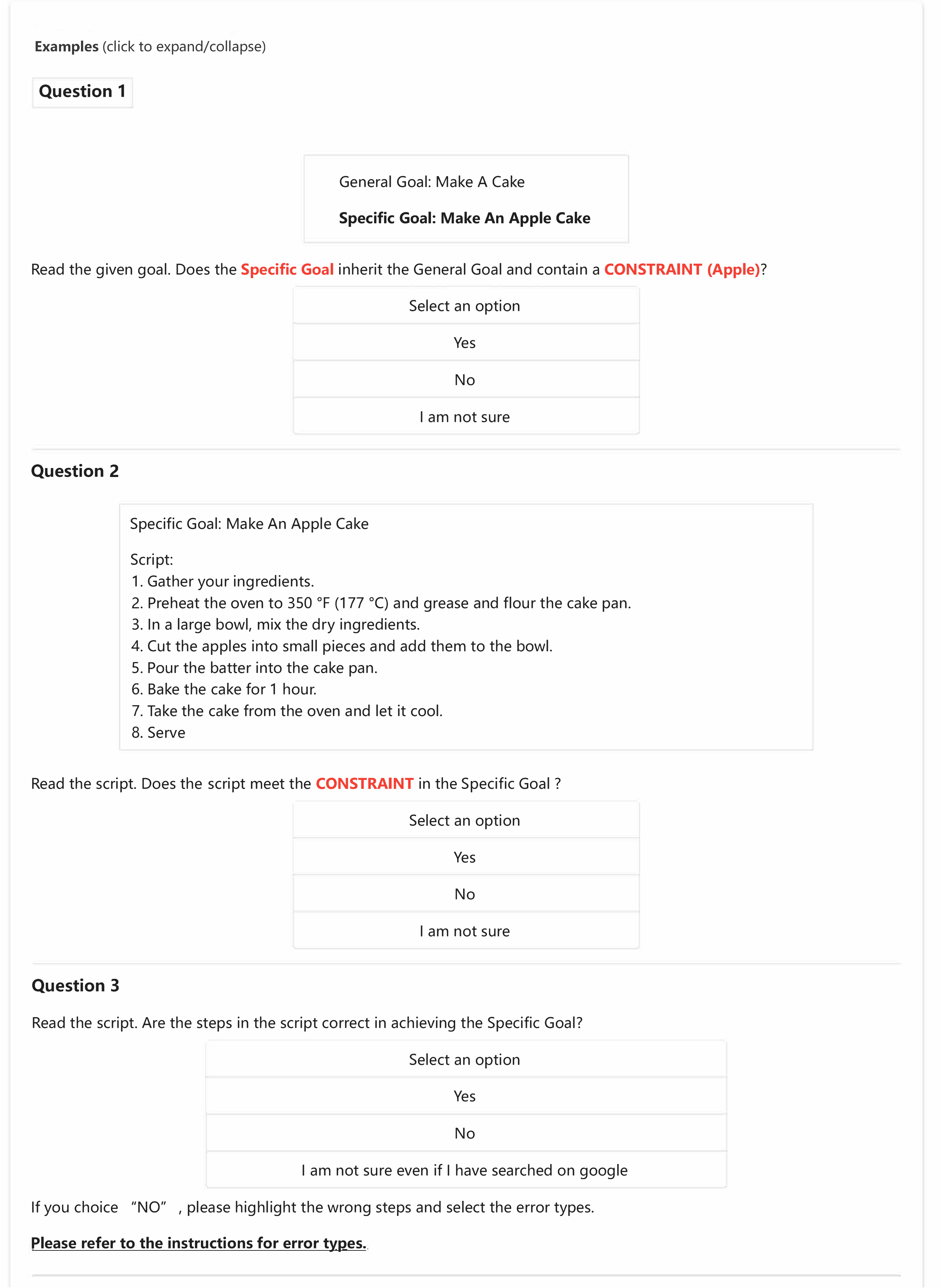}
    \caption{The examples given in the user interface.}
    \label{fig:interface}
\end{figure*}

\paragraph{Interface Details}
We conduct human evaluations on Amazon Mechanical Turk. 
Screenshots of the instructions and annotation interface are shown in Figure \ref{fig:instructions} and \ref{fig:interface}.

\paragraph{Step 1: Specific Goal Evaluation}
We first assess the specific goals generated by InstructGPT with three types of constraints.
We ask the turkers to check the specific goal whether inherits the abstract goal and contains a constraint.

\paragraph{Step 2: Script Evaluation}
In the second step, we show a script of the specific goal with actionable steps.
We then ask two questions:
\begin{enumerate}
    \item \textit{Does the script meet the constraint in the specific goal?} (Yes, No, or Not sure).  
    In our preliminary experiments, we found that the semantic completeness in the scripts generated based on InstructGPT (175B) is acceptable, but faithfulness to the constraints can not be guaranteed.
    This question assesses this tricky error;
    \item \textit{Are the steps in the script correct in achieving the specific goal?} (Yes, No, or Not sure).  
    This question is to assess whether the script can indeed accomplish the given goal.
    Although we have checked the constraint in the first question, there are still other error types (as shown in Figure~\ref{tab:type_error}).
    Then, we ask the turkers to review the generated scripts.
    If the scripts cannot achieve the given goal, they must point out the wrong steps and select the error types.
\end{enumerate}


\begin{table*}[htbp]
  \centering
  \small
  \resizebox{\textwidth}{!}
{
    \begin{tabular}{lll}
    \toprule
    \textbf{Task Prompt} & \multicolumn{2}{p{100mm}}{Create possible Specific Goals according to the Abstract Goal when the Constraint Type is XXX} \\
    \midrule
    \multicolumn{3}{c}{\textbf{Constraint Type: Modifier}} \\
    \midrule
    \makecell[l]{Example 1:\\Abstract Goal: Say Goodbye in Different Language\\Constraint: French\\Specific Goal: Say Goodbye in French\\Constraint: English\\Specific Goal: Say Goodbye in English} & 
    \makecell[l]{Example 2:\\Abstract Goal: Draw flowers\\Constraint: Pink\\Specific Goal: Draw pink flowers\\Constraint: Blue\\Specific Goal: Draw blue flowers} & 
    \makecell[l]{Example 3:\\Abstract Goal: Make hairstyle\\Constraint: At home\\Specific Goal: Make hairstyle at home\\Constraint: At a salon\\Specific Goal: Make hairstyle at a salon} \\
    \midrule
    \multicolumn{3}{c}{\textbf{Constraint Type: Method}} \\
    \midrule
    \makecell[l]{Example 1:\\Abstract Goal: Lower blood pressure\\Constraint: With medication\\Specific Goal: Lower blood pressure with medication\\Constraint: With exercises\\Specific Goal: Lower blood pressure with exercises} & 
   \makecell[l]{Example 2:\\Abstract Goal: Write a book\\Constraint: By hand\\Specific Goal: Write a book by hand\\Constraint: By typing\\Specific Goal: Write a book by typing} & 
    \makecell[l]{Example 3:\\Abstract Goal: Register to vote\\Constraint: Online\\Specific Goal: Register to vote online\\Constraint: Via mail\\Specific Goal: Register to vote via mail} \\
    \midrule
    \multicolumn{3}{c}{\textbf{Constraint Type: Intent}} \\
    \midrule
    \makecell[l]{Example 1:\\Abstract Goal: Make a cake\\Constraint: For a birthday party\\Specific Goal: Make a cake for a birthday party\\Constraint: For a weeding\\Specific Goal: Make a cake for a wedding} & 
    \makecell[l]{Example 2:\\Abstract Goal: Send an email\\Constraint: To get a gob\\Specific Goal: Send an email to get a gob\\Constraint: For leave\\Specific Goal: Send an email for leave}& 
    \makecell[l]{Example 3:\\Abstract Goal: Buy flowers\\Constraint: For girlfriend\\Specific Goal: Buy flowers for girlfriend\\Constraint: For mother\\Specific Goal: Buy flowers for mother} \\
    \bottomrule
    \end{tabular}%
    }
    \caption{The handpicked examples for specific goal generation.}
    \label{tab:handpicked}%
\end{table*}%

\begin{table*}[htbp]
  \centering
  \small
    \begin{tabular}{ll}
    \toprule
    \multicolumn{2}{c}{\textbf{Specific Goal: Wear a \colorbox[rgb]{0.99, 0.95, 0.93}{\color[rgb]{0.81, 0.41, 0.22}pencil} skirt}} \\
    \midrule
    \makecell[l]{Script (InstructGPT):\\
    1. Pick the right fabric.\\
    2. Choose the right length.\\
    3. Find the right fit.\\
    4. Accessorize appropriately.\\
    5. Wear the right shoes.
    } & \makecell[l]{Script (Ours):\\
    1. Find a pencil skirt that fits you well.\\
    2. Pair your pencil skirt with a fitted top.\\
    3. Choose the right shoes to complete your look.\\
    4. Accessorize as needed.
    \\ \\
    } \\
    \midrule
    \multicolumn{2}{c}{\textbf{Specific Goal: Clean your skin \colorbox[rgb]{0.99, 0.95, 0.93}{\color[rgb]{0.81, 0.41, 0.22}with soap and water}}} \\
    \midrule
    \makecell[l]{Script (InstructGPT):\\
    1. Get your skin wet with lukewarm water.\\
    2. Apply a small amount of cleanser to your skin.\\
    3. Rub the cleanser over your skin in gentle, circular motions.\\
    4. Rinse the cleanser off your skin with lukewarm water.\\
    5. Pat your skin dry with a clean, soft cloth.\\ \\ \\}
    & \makecell[l]{Script (Ours):\\
    1. Wet your skin with warm water.\\
    2. Apply a dollop of mild soap to your hands.\\
    3. Rub your hands together to create a lather.\\
    4. Apply the lather to your wet skin.\\
    5. Gently massage the lather into your skin in a circular motion.\\
    6. Rinse the soap off your skin with warm water.\\
    7. Pat your skin dry with a clean towel.} \\
    \midrule
    \multicolumn{2}{c}{\textbf{Specific Goal: Make stewed fruit \colorbox[rgb]{0.99, 0.95, 0.93}{\color[rgb]{0.81, 0.41, 0.22}with a crockpot}}} \\
    \midrule
    \makecell[l]{Script (InstructGPT):\\
    1. Combine all the ingredients in the crockpot.\\
    2. Cook on low for four hours.\\
    3. Serve with ice cream or yogurt.\\
    \\ \\ \\ 
    } & \makecell[l]{Script (Ours):\\
    1. Prepare the fruit.\\
    2. Add the sugar.\\
    3. Cook the fruit on low heat in the crockpot.\\
    4. Divide the fruit evenly into four bowls.\\
    5. Make your crumble.\\
    6. Sprinkle the crumble over your fruit and make it.} \\
    \bottomrule
    \end{tabular}%
    \caption{The generation comparison between InstructGPT (175B) with our method.
    We highlight the \colorbox[rgb]{0.99, 0.95, 0.93}{\color[rgb]{0.81, 0.41, 0.22}constraints} in the given examples.}
    \label{tb:case_study}%
\end{table*}%

\begin{table*}[htbp]
  \centering
  \small
    \begin{tabular}{ll}
    \toprule
    \textbf{Abstract Goal} & \multicolumn{1}{l}{\textbf{Constraints and Specific Goal}}\\
    \midrule
    Ask a teacher for help &\makecell[l]{Constraint: Math \\ Specific Goal: Ask a \colorbox[rgb]{0.99, 0.95, 0.93}{\color[rgb]{0.81, 0.41, 0.22}math} teacher for help \\ Constraint: Science \\ Specific Goal: Ask a \colorbox[rgb]{0.99, 0.95, 0.93}{\color[rgb]{0.81, 0.41, 0.22}science} teacher for help \\ Constraint: In school \\ Specific Goal: Ask a teacher for help \colorbox[rgb]{0.99, 0.95, 0.93}{\color[rgb]{0.81, 0.41, 0.22}in school}}\\ 
    \midrule
    Make pancakes & \makecell[l]{Constraint: Banana \\ Specific Goal: Make \colorbox[rgb]{0.99, 0.95, 0.93}{\color[rgb]{0.81, 0.41, 0.22}banana} pancakes \\ Constraint: Chocolate chip \\ Specific Goal: Make \colorbox[rgb]{0.99, 0.95, 0.93}{\color[rgb]{0.81, 0.41, 0.22}chocolate chip} pancakes}\\
    \midrule
    Download an xbox 360 game & \makecell[l]{Constraint: Halo 3 \\ Specific Goal: Download \colorbox[rgb]{0.99, 0.95, 0.93}{\color[rgb]{0.81, 0.41, 0.22}Halo 3} for Xbox 360 \\ Constraint: Gears of war \\ Specific Goal: Download \colorbox[rgb]{0.99, 0.95, 0.93}{\color[rgb]{0.81, 0.41, 0.22}Gears of War} for Xbox 360}\\
    \bottomrule
    \end{tabular}%
    \caption{Qualitative generations for specific goals when the type of constraints is \textit{Modifier}.}
    \label{table:PS_example_modifier}%
\end{table*}%

\begin{table*}[htbp]
  \centering
  \small
    \begin{tabular}{ll}
    \toprule
    \textbf{Abstract Goal} & \multicolumn{1}{l}{\textbf{Constraints and Specific Goal}}\\
    \midrule
    Prevent kidney stones from recurring & \makecell[l]{Constraint: By eating a healthy diet \\ Specific Goal: Prevent kidney stones from recurring \colorbox[rgb]{0.99, 0.95, 0.93}{\color[rgb]{0.81, 0.41, 0.22}by eating a healthy diet} \\ Constraint: By taking medication \\ Specific Goal: Prevent kidney stones from recurring \colorbox[rgb]{0.99, 0.95, 0.93}{\color[rgb]{0.81, 0.41, 0.22}by taking medication}}\\
    \midrule
    Sew chain stitch & \makecell[l]{Constraint: With a sewing machine \\ Specific Goal: Sew Chain Stitch \colorbox[rgb]{0.99, 0.95, 0.93}{\color[rgb]{0.81, 0.41, 0.22}with a sewing machine} \\ Constraint: By hand \\ Specific Goal: Sew chain stitch \colorbox[rgb]{0.99, 0.95, 0.93}{\color[rgb]{0.81, 0.41, 0.22}by jand}}\\
    \midrule
    Say goodbye in Spanish & \makecell[l]{Constraint: Formally \\ Specific Goal: Say goodbye in Spanish \colorbox[rgb]{0.99, 0.95, 0.93}{\color[rgb]{0.81, 0.41, 0.22}formally} \\ Constraint: Informally \\ Specific Goal: Say goodbye in Spanish \colorbox[rgb]{0.99, 0.95, 0.93}{\color[rgb]{0.81, 0.41, 0.22}informally}}\\
    \bottomrule
    \end{tabular}%
    \caption{Qualitative generations for specific goals when the type of constraints is \textit{Method}.}
    \label{table:PS_example_method}%
\end{table*}%

\begin{table*}[t]
\small
  \centering
    \begin{tabular}{ll}
    \toprule
    \textbf{Abstract Goal} & \multicolumn{1}{l}{\textbf{Constraints and Specific Goal}}\\
    \midrule
    Use clary sage & \makecell[l]{Constraint: For aromatherapy \\ Specific Goal: Use clary sage \colorbox[rgb]{0.99, 0.95, 0.93}{\color[rgb]{0.81, 0.41, 0.22}for aromatherapy} \\ Constraint: For skin care \\ Specific Goal: Use clary sage \colorbox[rgb]{0.99, 0.95, 0.93}{\color[rgb]{0.81, 0.41, 0.22}for skin care} \\ Constraint: For hair care \\ Specific Goal: Use clary sage \colorbox[rgb]{0.99, 0.95, 0.93}{\color[rgb]{0.81, 0.41, 0.22}for hair care}}\\
    \midrule
    Make carrot oil & \makecell[l]{Constraint: For skin \\ Specific Goal: Make carrot oil \colorbox[rgb]{0.99, 0.95, 0.93}{\color[rgb]{0.81, 0.41, 0.22}for skin} \\ Constraint: For cooking \\ Specific Goal: Make carrot oil \colorbox[rgb]{0.99, 0.95, 0.93}{\color[rgb]{0.81, 0.41, 0.22}for cooking}} \\
    \midrule
    Acquire abandoned property & \makecell[l]{Constraint: For personal use \\ Specific Goal: Acquire abandoned property \colorbox[rgb]{0.99, 0.95, 0.93}{\color[rgb]{0.81, 0.41, 0.22}for personal use} \\ Constraint: For business use \\ Specific Goal: Acquire abandoned property \colorbox[rgb]{0.99, 0.95, 0.93}{\color[rgb]{0.81, 0.41, 0.22}for business use}} \\
    \bottomrule
    \end{tabular}%
    \caption{Qualitative generations for specific goals when the type of constraints is \textit{Intent}.}
    \label{table:PS_example_intent}%
\end{table*}%

\begin{table*}[htbp]
  \centering
  \small
    \begin{tabular}{ll}
    \toprule
    \multicolumn{2}{c}{\textbf{Abstract Goal: Link social media accounts on filpboard }} \\
    \midrule
    \makecell[l]{Specific Goal: \\ Link \colorbox[rgb]{0.99, 0.95, 0.93}{\color[rgb]{0.81, 0.41, 0.22}facebook} accounts on flipboard} & \makecell[l]{ Specific Goal: \\ Link \colorbox[rgb]{0.99, 0.95, 0.93}{\color[rgb]{0.81, 0.41, 0.22}twitter} accounts on flipboard}\\
    \midrule
    \makecell[l]{Script:\\ 1. Launch Flipboard. \\ 2. Go to Content. \\ 3. Go to Accounts. \\ 4. Select Facebook. \\ \\} & \makecell[l]{Script:\\1. Launch Flipboard. \\ 2. Go to Content. \\ 3. Go to Accounts. \\ 4. Select Twitter. \\ 5. Log into your Twitter account.} \\
    \midrule
    \multicolumn{2}{c}{\textbf{Abstract Goal: Increase your metabolism }} \\
    \midrule
    \makecell[l]{Specific Goal: \\ Increase your metabolism \colorbox[rgb]{0.99, 0.95, 0.93}{\color[rgb]{0.81, 0.41, 0.22}with medication}} & \makecell[l]{Specific Goal: \\ Increase your metabolism \colorbox[rgb]{0.99, 0.95, 0.93}{\color[rgb]{0.81, 0.41, 0.22}with diet}} \\
    \midrule
    \makecell[l]{Script:\\1. Talk to your doctor about prescription options. \\ 2. Follow your doctor's instructions carefully. \\ 3. Stay hydrated } & \makecell[l]{Script:\\1. Eat small, frequent meals throughout the day. \\ 2. Choose lean proteins. \\ 3. Add spice to your favorite foods.} \\
    \midrule
    \multicolumn{2}{c}{\textbf{Abstract Goal: Use clary sage }} \\
    \midrule
    \makecell[l]{Specific Goal: \\ Use clary sage \colorbox[rgb]{0.99, 0.95, 0.93}{\color[rgb]{0.81, 0.41, 0.22}for aromatherapy}} & \makecell[l]{Specific Goal: \\ Use clary sage \colorbox[rgb]{0.99, 0.95, 0.93}{\color[rgb]{0.81, 0.41, 0.22}for massage}} \\
    \midrule
    \makecell[l]{Script:\\1. Make a clary sage and lavender sachet to keep in your pillowcase. \\ 2. Diffuse clary sage oil to boost your mood and energy levels. \\ 3. Add a few drops of clary sage oil to your bathtub for a relaxing soak \\ \\} & \makecell[l]{Script:\\1. Heat clary sage oil until it is warm. \\ 2. Massage the oil into the muscles and tissues. \\ 3. Leave the oil on for at least 30 minutes. \\ 4. Rinse the oil off with warm water.} \\
    \bottomrule
    \end{tabular}%
    \caption{Qualitative generations for scripts under the different \colorbox[rgb]{0.99, 0.95, 0.93}{\color[rgb]{0.81, 0.41, 0.22}constraints}.}
    \label{table:GE_example}%
\end{table*}%

\label{sec:appendix}
\end{document}